\documentclass[10pt,twocolumn,letterpaper]{article}

\usepackage[pagenumbers]{cvpr} 

\usepackage{graphicx}
\usepackage{amsmath}
\usepackage{amssymb}
\usepackage{booktabs}
\usepackage{enumitem}

\usepackage{caption}
\usepackage[pagebackref,breaklinks,colorlinks]{hyperref}

\usepackage[capitalize]{cleveref}
\crefname{section}{Sec.}{Secs.}
\Crefname{section}{Section}{Sections}
\Crefname{table}{Table}{Tables}
\crefname{table}{Tab.}{Tabs.}

\def\cvprPaperID{5874} 
\def\confName{CVPR}
\def\confYear{2022}

\begin{document}

\title{Smooth-Swap: A Simple Enhancement for Face-Swapping with Smoothness}

\author{
Jiseob Kim\textsuperscript{1,2}, Jihoon Lee\textsuperscript{2}, Byoung-Tak Zhang\textsuperscript{1}\\
\textsuperscript{1}Seoul National University, 
\textsuperscript{2}Kakao Brain\\
{\tt\small jkim@bi.snu.ac.kr, jihoonlee.in@gmail.com, btzhang@bi.snu.ac.kr}
}


\twocolumn[{%
\renewcommand\twocolumn[1][]{#1}%
\maketitle
\begin{center}
    \centering
    \includegraphics[width=1.0\textwidth]{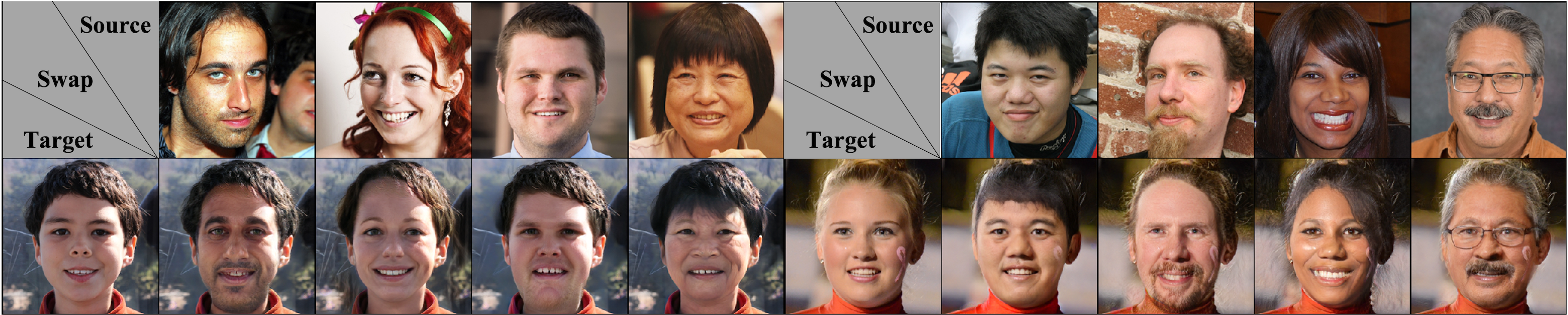}
    \captionof{figure}{Face-swapped images generated by our \textit{Smooth-Swap} model. In the swapped images, the identities of the target images are replaced with that of the source images. See the face-shape, hair, and mustache change in accordance with different sources.}
\end{center}%
}]


\begin{abstract}

Face-swapping models have been drawing attention for their compelling generation quality, but their complex architectures and loss functions often require careful tuning for successful training. We propose a new face-swapping model called `Smooth-Swap', which excludes complex handcrafted designs and allows fast and stable training. The main idea of Smooth-Swap is to build smooth identity embedding that can provide stable gradients for identity change. Unlike the one used in previous models trained for a purely discriminative task, the proposed embedding is trained with a supervised contrastive loss promoting a smoother space. With improved smoothness, Smooth-Swap suffices to be composed of a generic U-Net-based generator and three basic loss functions, a far simpler design compared with the previous models. Extensive experiments on face-swapping benchmarks (FFHQ, FaceForensics++) and face images in the wild show that our model is also quantitatively and qualitatively comparable or even superior to the existing methods.

\end{abstract}


\section{Introduction}

Face swapping is a task to switch the person-identity of a given face image with another, preserving other attributes like facial expressions, head poses, and backgrounds. The task has been highlighted for its wide use of real-world applications, such as anonymization in privacy protection and the creation of new characters in the entertainment industry. With progress made over years \cite{baoOpenSetIdentityPreserving2018b,natsumeRSGANFaceSwapping2018,nirkinFSGANSubjectAgnostic2019,Li_2020_CVPR,chenSimSwapEfficientFramework2020,thiesDeferredNeuralRendering2019,zhuOneShotFace2021,wangHifiFace3DShape2021}, state-of-the-art face-swapping models can generate a swapped image of decent quality using a single shot of a new source identity. 

Despite the performance improvement, however, existing models usually adopt complex model architectures and numerous loss functions to change \emph{face shape}. Face shape is a crucial component of identity, but changing it is a nontrivial task; it incurs a dramatic change of pixels, but no guidance can be given due to the inherent absence of the ground-truth swapped images. Thus, previous studies have focused on using handcrafted components such as mask-based mixing~\cite{chenSimSwapEfficientFramework2020} or 3D face-shape modeling~\cite{Li_2020_CVPR,wangHifiFace3DShape2021}. Although such components are effective for changing shape and improving the swapped-image quality, the models have added complexity of hyperparameters and loss functions that require careful tuning for successful training.

In this study, we postulate that the approaches based on handcrafted components are not the best way to resolve the difficulty of face-swapping. We propose instead a new identity embedding model having improved smoothness, which we assume to be related most to the gist of the problem. An identity embedding model, or an embedder, plays a key role during the training of the swapping model. It gives gradients for the generator, to which direction it has to tune to change the identity. It is thus important the embedder has a smooth space, since the gradients can be erroneous or noisy otherwise. In our proposed model, \textit{Smooth-Swap}, we consider a new embedder trained with supervised constrastive loss \cite{khoslaSupervisedContrastiveLearning2020}. \cite{wangUnderstandingContrastiveRepresentation2020}. We find it has a smoother space than the ArcFace embedder \cite{dengArcFaceAdditiveAngular2019}, one used in the most of the existing models, and helps faster and stable training.

\begin{figure*}
    \centering
    \includegraphics[width=\textwidth]{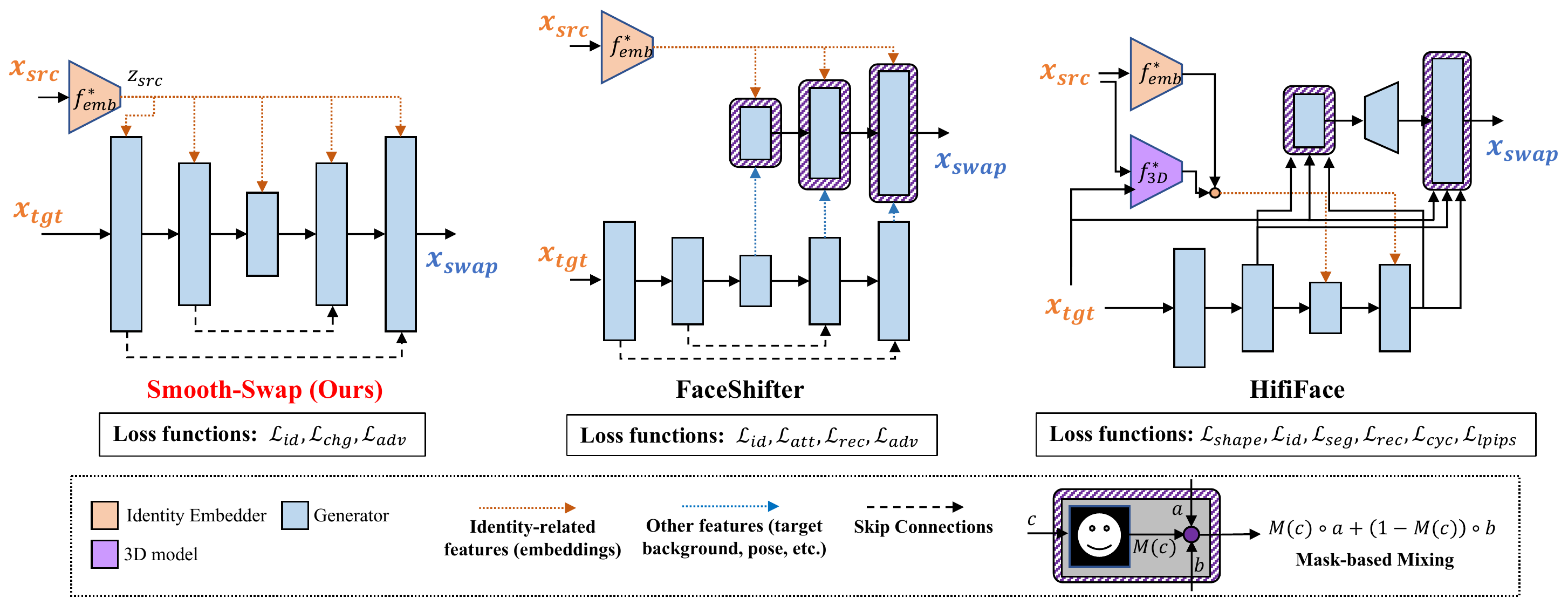}
    \caption{An illustrative comparison of the generator architectures and the loss functions of face-swapping models. Previous models (FaceShifter \cite{Li_2020_CVPR} and HifiFace \cite{wangHifiFace3DShape2021}) have face-swapping-specific designs such as mask-based mixing (hatched in purple) or 3D face modeling ($f_{3D}^*$). Such designs induce complex architectures and various loss functions, which makes training difficult for balancing. On the contrary, our architecture is a simple U-Net extension excluding task-related heuristics, and trained by only three typical losses.}
    \label{fig:arch_compare}
\end{figure*}

Through the smooth embedder, Smooth-Swap works without any handcrafted components. It adopts a simple U-Net \cite{ronnebergerUNetConvolutionalNetworks2015}-based generator, and we train it using only three basic loss functions---identity change, target preserving, and adversarial (Fig. \ref{fig:arch_compare}). While this set-up is simpler than the existing models, we find that our model can still achieve comparable or superior performance by taking a data-driven approach and minimizing inductive bias.

The advantages of Smooth-Swap can be summarized as follows. \textbf{1) Simple architecture}: 
Smooth-Swap uses a simple U-Net \cite{ronnebergerUNetConvolutionalNetworks2015}-based generator, which does not involve any handcrafted components as the existing models.
\textbf{2) Simple loss functions}: The Smooth-Swap generator can be trained using minimal loss functions for face-swapping---identity, pixel-level change, and adversarial loss.
\textbf{3) Fast training}: The smooth identity embedder allows faster training of the generator by providing more stable gradient information.

\section{Related Work}

\paragraph{Approaches based on 3D Models and Segmentation}
Earlier face-swapping models rely on external modules such as 3D Morphable Models (3DMM) \cite{blanzMorphableModelSynthesis1999} and a face segmentation model. Face2Face~\cite{thiesFace2FaceRealtimeFace2016} and \cite{nirkinFaceSegmentationFace2018} fit the source and the target images to 3DMM and transfers the expression (and the posture) parameters to synthesize the swapped image. 
RSGAN \cite{natsumeRSGANFaceSwapping2018}, FSNet \cite{natsumeFSNetIdentityAwareGenerative2018}, and FSGAN \cite{nirkinFSGANSubjectAgnostic2019} use a segmentation model to separate the facial region from the background, generate the swapped image by switching and blending the regions.
Despite the early success, these approaches do not produce high quality images since their performance depends on the non-trainable external modules.

\paragraph{Feature-based GAN models}
In contrast with the approaches above, recent models consider end-to-end training, generating a face-swapped image based on learned features.
IPGAN \cite{baoOpenSetIdentityPreserving2018b} learns separate embedding vectors for the identity and the target attributes, switching and recombining them to generate a swapped image.
FaceShifter \cite{Li_2020_CVPR} considers multi-level mixing using an encoder-decoder architecture, alleviating the information loss in the approach of IPGAN.
SimSwap~\cite{chenSimSwapEfficientFramework2020} proposes weak feature matching to focus more on preserving the facial expression of the source, whereas HifiFace \cite{wangHifiFace3DShape2021} proposes a method integrating 3D shape model to focus more on active shape change. InfoSwap \cite{gaoInformationBottleneckDisentanglement2021} uses information bottleneck for better disentangling the identity attributes from the rest.  
MegaFS~\cite{zhuOneShotFace2021} utilizes a pretrained StyleGAN2~\cite{karrasAnalyzingImprovingImage2020} to generate high-resolution face-swapped images. \cite{naruniecHighResolutionNeuralFace2020} also tackles high resolution by training a separate generator for each identity.
Although these models have continuously improved the quality of the generated images, they tend to show weak identity change or involve complexity due to handcrafted components.

\section{Problem Formulation \& Challenges} 
\label{sec:prob_def}

We first describe the problem formulation and main technical challenges of face-swapping. Then, we introduce how the smoothness of an identity embedder can alleviate them.

\subsection{Problem Formulation}

When a source $x_{src}$ and a target $x_{tgt}$ are given, a face-swapping model needs to generate the swap image, $x_{swap}$, which satisfies the following conditions:
\begin{enumerate}[label={C\arabic*.}]
    \item It has the identity of the source image.
    \item Other than the identity, it looks the same as the target image (having the same background, pose, etc.).
    \item It looks realistic (indistinguishable from real images).
\end{enumerate}
To meet these requirements, most of face-swapping models~\cite{Li_2020_CVPR,wangHifiFace3DShape2021} consist of three components: an identity embedder $f_{emb}^*$ for the source image, a generator $f_{gen}$ for the swapped image, and a discriminator $f_{dis}$ to improve the fidelity. Fig.~\ref{fig:arch_compare} shows an overview of these face-swapping models including our approach. 
Note that the identity embedder is pre-trained and frozen during the training of other components, so the asterisk is included in the superscript. 

\subsection{Challenges for Changing Identity}

\begin{figure}[ht]
    \centering
    \includegraphics[width=\columnwidth]{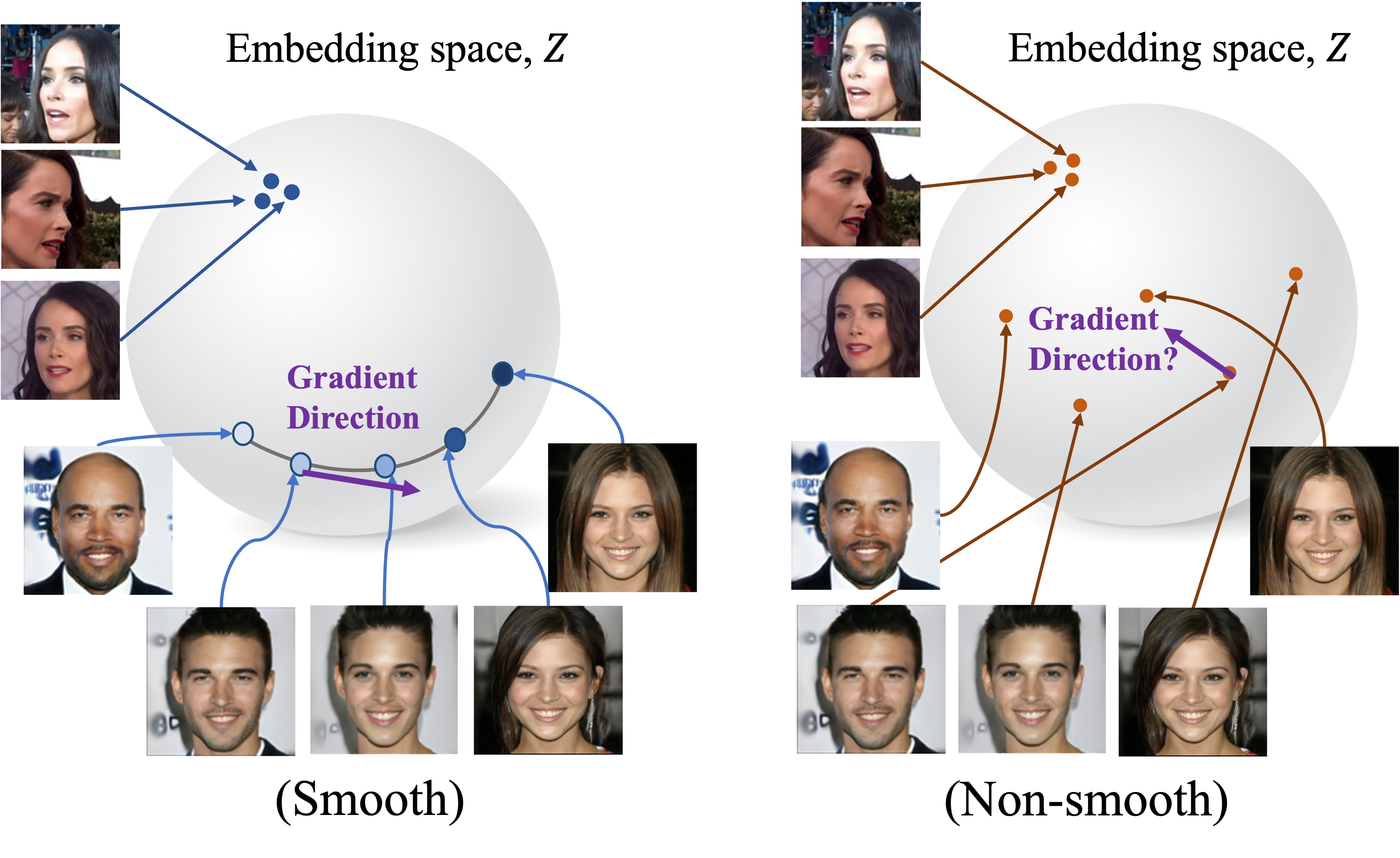}
    \vspace{-20px}
    \caption{When identity is changed from one to another, the corresponding vector in a smooth embedding space would also change smoothly. In a non-smooth embedding space, however, the vector would make discrete jumps. The space can become non-smooth if the embedder is strongly trained on a discriminative task. In this case, the embedder cannot give a good gradient direction for the generator to change the identity correctly. See \ref{sec:imp_emb}.}
    \label{fig:why_smooth_emb}
\end{figure}

The main difficulty for training a face-swapping model comes from the conflict between C1 and C2. Satisfying C1 makes $x_{swap}$ move away from $x_{tgt}$ to change the identity, whereas satisfying C2 enforces it to stay around. If we can accurately extract the \textit{identity-irrelevant} change of $x_{swap}$ from $x_{tgt}$ and use it for the loss of C2, this conflict would have been relaxed. Unfortunately, designing such a loss is difficult, and a common fallback is to use an isotropic loss such as perceptual \cite{zhangUnreasonableEffectivenessDeep2018} or pixel-level $L_p$ loss. 

A major consequence of the conflict and an isotropic C2 loss is stagnant face-shape change. Shape-wise change such as round to sharp chin involves geometric transformations and entails dramatic variation in features and pixel values. It is thus a big fight against the C2 loss preventing any aspects of deviation from $x_{tgt}$ and often compromised first. In this regard, previous work put much effort for changing face shape correctly, using a 3D face model, for example, to better capture the shape \cite{wangHifiFace3DShape2021}. However, such a design introduces additional complication and requires a careful balancing between modules for successful training. In this work, we hypothesize that the conflict can be relaxed not by adding new modules but by introducing smoothness to an identity embedder. We will describe the details on this in the following section. 

\subsection{Importance of A Smooth Identity Embedder} \label{sec:imp_emb}

Most of the previous face-swapping models use ArcFace~\cite{dengArcFaceAdditiveAngular2019} as an identity embedder (embedder for short) since it is one of the state-of-the-art face recognition models. Feeding images into the embedder and comparing features from the last layer (called embedding vectors), it provides a decent similarity metric for the person-identities of face images. Using ArcFace or any other face recognition models, we typically deal with a highly non-smooth embedding space, because these are trained only by a discriminative task. 

The smoothness of the embedder, however, is crucial during the training of a face-swapping model. When a model generates $x_{swap}$ with a wrong identity amid training, the embedder has to give a good gradient direction to correct it. This gradient has to be accurate and consistent; otherwise $x_{swap}$ easily goes back to $x_{tgt}$ by the loss for C2. If the embedding space is non-smooth, the gradient direction can be erroneous or noisy since gradients are only well-defined in a continuous space.

\section{Method: Smooth-Swap}
We explain our main model called \textit{Smooth-Swap}. The model introduces a new identity embedder, trained using supervised contrastive learning \cite{khoslaSupervisedContrastiveLearning2020} to improve the smoothness in the embedding space. It also introduces a simple U-Net style generator architecture, which is well suited to the new identity embedder.  

\paragraph{Notations} 
Our identity embedder takes images $x\in X$ and outputs the corresponding embedding vectors $z\in Z$ (e.g., $z_{src}:=f_{emb}^*(x_{src})$). The generator takes a target image $x_{tgt}$ and a source embedding $z_{src}$, and produces the swap image: $x_{swap}=f_{gen}(x_{tgt}, z_{src})$. $f_{dis}$ takes $x_{swap}$ and outputs a scalar ranging $[0,1]$ (close to $0$ for fake and $1$ for real).

\subsection{Smooth Identity Embedder} \label{sec:smooth_emb}

As discussed in Sec. \ref{sec:imp_emb}, we desire a smooth embedder for stable and effective training. To train such an embedder, we consult a supervised contrastive learning loss \cite{khoslaSupervisedContrastiveLearning2020}:
\[
\mathcal{L}(f_{emb})=\underset{(x_{i},x_{p}^i,x_n^i)
}{\mathbb{E}}\left[-\log\frac{e^{\left(\left\langle z_{i},z_{p}\right\rangle /\tau\right)}}{e^{\left(\left\langle z_{i},z_{p}\right\rangle /\tau\right)}+\sum_{n}e^{\left(\left\langle z_{i},z_{n}\right\rangle /\tau\right)}}\right]
\]
where $x_i$ denotes a sample from the training dataset; $x_p^i$ and $x_n^i$ denote positive (images having the same identity as $x_i$) and negative (having a different identity) samples, respectively.

An important property of contrastive learning is that it makes the embedding vectors keep the maximal information \cite{wangUnderstandingContrastiveRepresentation2020}, and this is closely related to our need of a smooth embedder. If we have face images of the same identity but of a different age or of a different face shape (e.g., from a diet), discriminative embedders like ArcFace \cite{dengArcFaceAdditiveAngular2019} remove this information aggressively to align the embedding vectors. While this is beneficial for classifying identities, it incurs a non-smooth embedding space. When changing the identity from elderly to young or from a round shape to sharp in this space, the embedding vectors cannot change smoothly as such information is removed. For our purpose, more desired are the embeddings with richer information---even if the alignment is compromised---as can be obtained from the contrastive learning. Then, changing from one identity to another is a smooth path and a good gradient direction can be obtained for training the swapping model (see Fig. \ref{fig:why_smooth_emb}).

\subsection{Generator Architecture} \label{sec:arch}

Our generator architecture is an adaptation from the noise conditional score network (NCSN++), which is one of the state-of-the-art architectures in score-based generative modeling \cite{songScoreBasedGenerativeModeling2021} (Fig. \ref{fig:arch_compare}). While the original usage of NCSN++ is far different from face-swapping, we find its U-Net nature \cite{ronnebergerUNetConvolutionalNetworks2015} and conditioning structure is useful for our task. We modify two parts from NCSN++; the time embedding is replaced with the identity embedding and a direct skip connection from the input to the output is added. 

\paragraph{Details on Structure} NCSN++ is basically a U-Net \cite{ronnebergerUNetConvolutionalNetworks2015} with a conditioning structure and modern layer designs such as residual and attention blocks. Its original goal is to take a noisy image and output a score vector having the same dimensionality as the image. Since it has to output a vector conditioned on varying noise levels controlled by time, it also takes a time embedding vector that is added to each residual block after being broadcasted over the width and height dimensions. In our design, we replace this embedding vector with identity embedding, as illustrated in Fig. \ref{fig:arch_compare}. Also, since the score vector is close to a difference between images rather than an image itself, we add the input image when making the final output image, instead of directly passing the output (i.e., an input-to-output skip connection). 

Note our architecture does not include any task-specific design components such as a 3D face model or mask-based mixing from the previous work. It is universal and mostly compatible with score modeling by design.

\paragraph{Loss Functions}
\label{sec:loss}
To train this generator, we use three most basic loss functions, each corresponding to the conditions for $x_{swap}$ described at the beginning of Sec. \ref{sec:prob_def}. 

\[
\begin{cases}
\mathcal{L}_{id}&=~1-\cos(z_{swap}, z_{src})\\
\mathcal{L}_{chg}&=~\Vert x_{swap} - x_{tgt} \Vert_2^2 / D \\
\mathcal{L}_{adv}&=~-\log (f_{dis}(x_{swap}))
\end{cases}
\]
The total loss is computed by combining these functions and taking the expectation over ($x_{tgt}$, $x_{src}$) pairs:
\[
    \mathcal{L}(f_{gen}) = \mathbb{E}_{(x_{tgt}, x_{src})} [\lambda_{id} \mathcal{L}_{id} + \lambda_{chg} \mathcal{L}_{chg} + \lambda_{adv} \mathcal{L}_{adv}].
\]
Note that $\cos(\cdot,\cdot)$ stands for cosine similarity and $D$ stands for the number of dimensions of $X$; $f_{dis}$ is trained with the original loss from  \cite{goodfellowGenerativeAdversarialNetworks2014} and R1 regularizer \cite{meschederWhichTrainingMethods2018a}. The loss functions are generally the same as \cite{Li_2020_CVPR}, except we use a simpler pixel-level change loss instead of the feature-level loss (denoted as attribute loss in the paper). For each minibatch, we include one ($x_{tgt}$, $x_{tgt}$) pair, whose change loss effectively acts as a reconstruction loss.

\section{Experiments}\label{sec:exp}

\begin{table*}[ht]
    \centering
\resizebox{\textwidth}{!}{
{\small
\begin{tabular}{lrccccc|ccccc|c}
\toprule

\textbf{Model}           &   \textbf{VGG↓} &   \textbf{VGG-R↓} &   \textbf{Arc↑} &   \textbf{Arc-R↑} &   \textbf{Shp↓} &   \textbf{Shp-R↓} &   \textbf{Expr↓} &   \textbf{Expr-R↓} &   \textbf{Pose↓} &   \textbf{Pose-R↓} &   \textbf{PoseHN↓} & \textbf{Overall↓}\\

\midrule 
Deepfakes  & 120.907  & 0.493  & 0.443  & 0.524  & 0.639  & \textbf{0.464 } & 0.802  & 0.541  & 0.188  & 0.445  & 4.588 & 0.927\tabularnewline
FaceShifter  & 110.875  & 0.482  & $\dagger$ & $\dagger$ & 0.658  & 0.492  & \textbf{0.653 } & 0.456  & \textbf{0.177 } & \textbf{0.381 } & \textbf{3.175 } & -0.202\tabularnewline
SimSwap  & \textbf{99.736 } & \textbf{0.435 } & $\dagger$ & $\dagger$ & 0.662  & 0.479  & \textbf{0.644 } & \textbf{0.449 } & 0.178  & \textbf{0.385 } & 3.749 & \textbf{-0.558}\tabularnewline
HifiFace  & 106.655  & 0.469  & \textbf{0.527 } & \textbf{0.550 } & \textbf{0.616}  & 0.465  & 0.702  & 0.484  & \textbf{0.177 } & 0.387  & \textbf{3.370 } & -0.329\tabularnewline
MegaFS  & 110.897  & 0.461  & $\dagger$ & $\dagger$ & 0.701  & 0.500  & 0.678  & \textbf{0.436 } & 0.182  & 0.398  & 5.456 & 0.234\tabularnewline
\midrule
Smooth-Swap  & \textbf{101.678 } & \textbf{0.435 } & \textbf{0.464 } & \textbf{0.611 } & \textbf{0.565 } & \textbf{0.403 } & 0.722  & 0.477  & 0.186  & 0.395  & 4.498 & \textbf{-0.617}\tabularnewline
\midrule
\midrule
50\% steps & 101.905 & 0.430 & - & - & 0.578 & 0.404 & 0.726 & 0.476 & 0.186 & 0.399 & 5.979 & -0.398\tabularnewline
$\lambda_{id}=1$  & 107.096  & 0.446  & 0.421  & 0.581  & 0.610  & 0.415  & 0.669  & 0.461  & 0.185  & 0.398  & 4.636 & -0.419\tabularnewline
(Arc) $\lambda_{id}=1$  & 103.767  & 0.437  & $\dagger$  & $\dagger$  & 0.682  & 0.460  & 0.728  & 0.493  & 0.192  & 0.416  & 5.457 & 0.266\tabularnewline
(Arc) $\lambda_{id}=4$  & 98.115  & 0.421  & $\dagger$  & $\dagger$  & 0.684  & 0.441  & 0.914  & 0.543  & 0.207  & 0.430  & 5.655 & 0.699\tabularnewline
\bottomrule
\end{tabular}
}} 

\caption*{\scriptsize{\textbf{Shp}: shape, \textbf{Expr}: expression, \textbf{PoseHN}: pose metric with Hopenet \cite{sanyalLearningRegress3D2019}, (Arc): trained using ArcFace, $\dagger$: scores cannot be compared because the model uses ArcFace in training.}}

\caption{Quantitative comparison between the models (see Sec. \ref{sec:eval-detail} and Sec. \ref{sec:basic-performance} for the details). The arrow ↓ (or ↑) denotes that the score is the lower (or the higher) the better; the best two are marked as bold. The vertical line in the middle divides the scores into two groups: ones related to the identity change (left) and ones related to keeping the target attributes (right). The overall score is the average of each score after standardization (Arc and Arc-R are excluded as some models are ineligible). The last four rows are ablation models (Sec. \ref{sec:ablation}).}

\label{tab:synth}
\end{table*}

\subsection{Training Details}
\paragraph{Datasets}

For training the generator, we use FFHQ dataset \cite{karrasStyleBasedGeneratorArchitecture2019}, which contains 70k aligned face images. We use the 10\% of images for testing. For training the identity embedder, we use the VGGFace2 dataset \cite{caoVGGFace2DatasetRecognising2018}, which contains 3.3M identity-labeled images of 9k subjects. We crop and align VGGFace2 images using the same procedure as FFHQ. All images including FFHQ are resized to 256$\times$256 scale.

\paragraph{Architecture Details}

Our identity embedder is based on ResNet50 \cite{heDeepResidualLearning2016} architecture. The final, average-pooled feature vector is passed through two fully-connected layers and normalized to unit length. The generator architecture is mostly the same as NCSN++ \cite{songScoreBasedGenerativeModeling2021}, except we use half as many channels. The discriminator is set to the same as StyleGAN2 \cite{karrasAnalyzingImprovingImage2020}. The detailed structure of the networks is included in the appendix. 

\paragraph{Training}
We set $\lambda_{id}=4$, $\lambda_{chg}=1$, and $\lambda_{adv}=1$ for training. The discriminator is trained with the non-saturating loss \cite{goodfellowGenerativeAdversarialNetworks2014} along with the R1 regularizer \cite{meschederWhichTrainingMethods2018a} to prevent the overfitting.  
Adam optimizer \cite{kingmaAdamMethodStochastic2015} is used for training with learning rates 0.001 (generator) and 0.004 (discriminator). It is run for 800k steps with batch-size eight, where the number matches with the total number of images shown to HifiFace. As described in Sec. \ref{sec:loss}, one pair in the batch is set to ($x_{tgt}$, $x_{tgt}$) for considering the self-reconstruction case. 
Adam is also used for training the embedder (prior to training the swapping model), where the learning rate is set to 0.001 and decreased by a factor of 10 at 60, 75, and 90\% during the total 101K steps. The batch size is 128 (32 identities, four instances per each) and the temperature $\tau$ is 0.07 as suggested in \cite{khoslaSupervisedContrastiveLearning2020}.

\subsection{Evaluation Details} \label{sec:eval-detail}

\paragraph{Compared Models}
We compare our Smooth-Swap model with the latest feature-based face-swapping models: FaceShifter \cite{Li_2020_CVPR}, MegaFS \cite{zhuOneShotFace2021},  HifiFace \cite{wangHifiFace3DShape2021}, SimSwap \cite{chenSimSwapEfficientFramework2020}, and Neural Textures \cite{thiesDeferredNeuralRendering2019}.
We also compare two of the earliest models: Deepfakes \cite{DeepFakesHttpsGithub2021} and Faceswap \cite{FaceSwapHttpsGithub2021}.

\paragraph{Quantitative Evaluations}

Since the most of the compared models do not open their source code to the public, the current standard for evaluating the models is to compare their generated images\footnote{Available on \url{https://github.com/ondyari/FaceForensics}; some are on the project page of each model.} on the FaceForensics++ (FF++) datasets \cite{rosslerFaceForensicsLearningDetect2019}, and we follow accordingly.

We evaluate various metrics that can be grouped into the following: identity, shape, expression, and pose. We want $x_{swap}$ to be close to $x_{src}$ for the first two and close to $x_{tgt}$ for the other two. To evaluate identity, we use VGGFace2 \cite{caoVGGFace2DatasetRecognising2018} and ArcFace \cite{dengArcFaceAdditiveAngular2019} embedders and compute the embedding distance and cosine similarity, respectively, between $x_{swap}$ and $x_{src}$. Compared with the retrieval accuracy used in \cite{wangHifiFace3DShape2021, Li_2020_CVPR, chenSimSwapEfficientFramework2020}, which classifies $x_{swap}$ among fixed candidates, this metric allows more fine-grained comparison. To evaluate shape, expression, and pose, we follow the evaluation protocol of \cite{wangHifiFace3DShape2021}; i.e., we use a 3D face model of \cite{sanyalLearningRegress3D2019} to get the parameters of each class and compute the L2 distances.

When applicable, we compute relative distances and similarities (denoted by '-R') as well. For example, 
$$
\text{dist-R} := \frac{\text{dist}(x_{swap},x_{src})}{\text{dist}(x_{swap},x_{src})+\text{dist}(x_{swap},x_{tgt})},
$$
is computed for VGGFace2 embedding distance\footnote{For pose and expressions, numerator is changed to $(x_{swap},x_{tgt})$}. This is to reflect how humans perceive the changes; to our eyes, important is not only the identity of $x_{swap}$ being close to $x_{src}$ but also its being far from $x_{tgt}$.

\begin{figure*}[ht]
    \centering
    \includegraphics[width=1.0\textwidth]{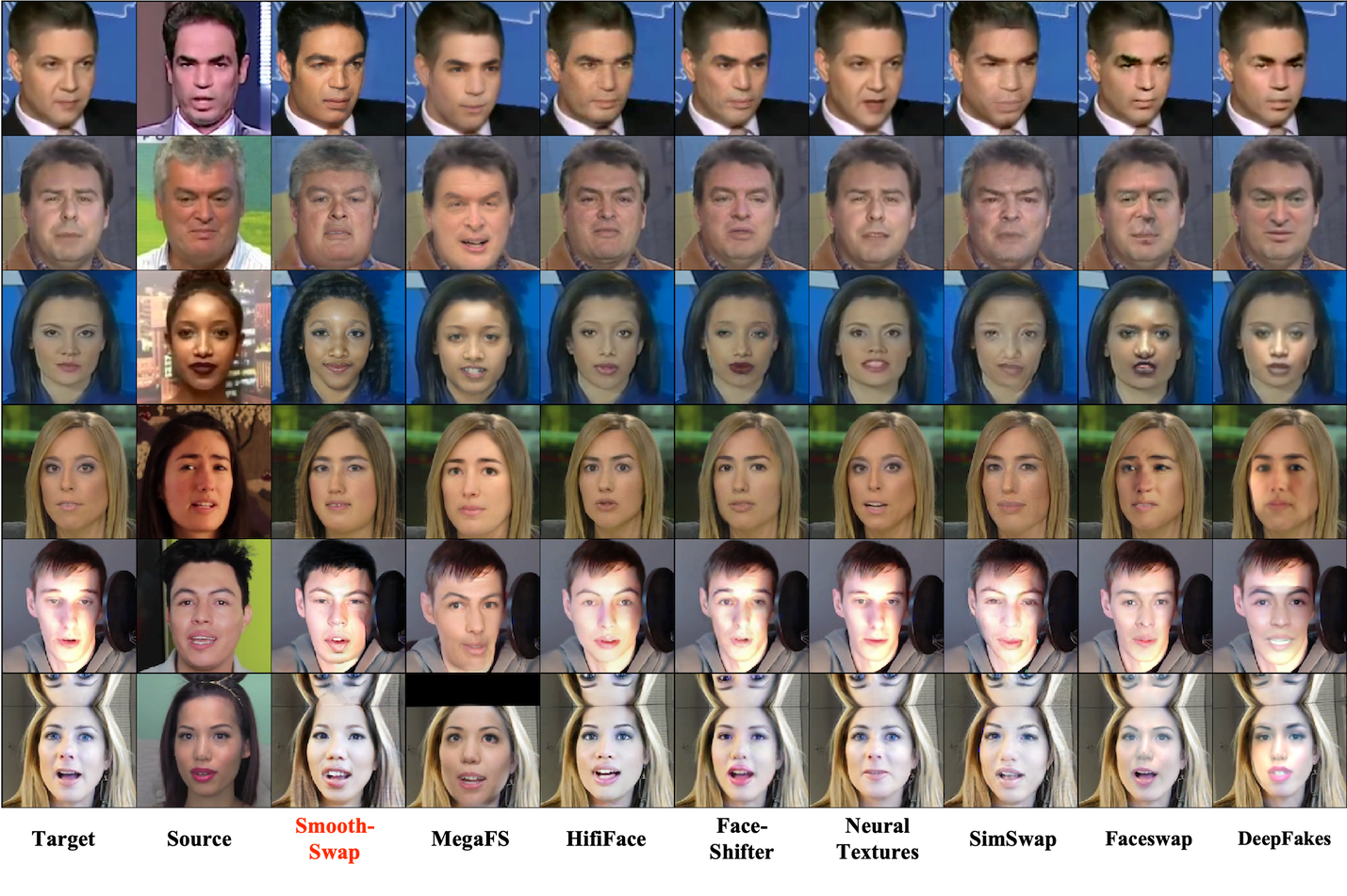}
    \caption{Comparison of the face-swapping results of various models on the FaceForensics++ dataset \cite{rosslerFaceForensicsLearningDetect2019}. The results from our models show the most active identity and shape change, reflecting the characteristics of the source identities. Note there are minor frame differences among the results as the images are extracted from videos.}
    \label{fig:compare}
\end{figure*}

\subsection{Basic Face-Swapping Performance} \label{sec:basic-performance}
\begin{figure}
    \centering
    \includegraphics[width=\columnwidth]{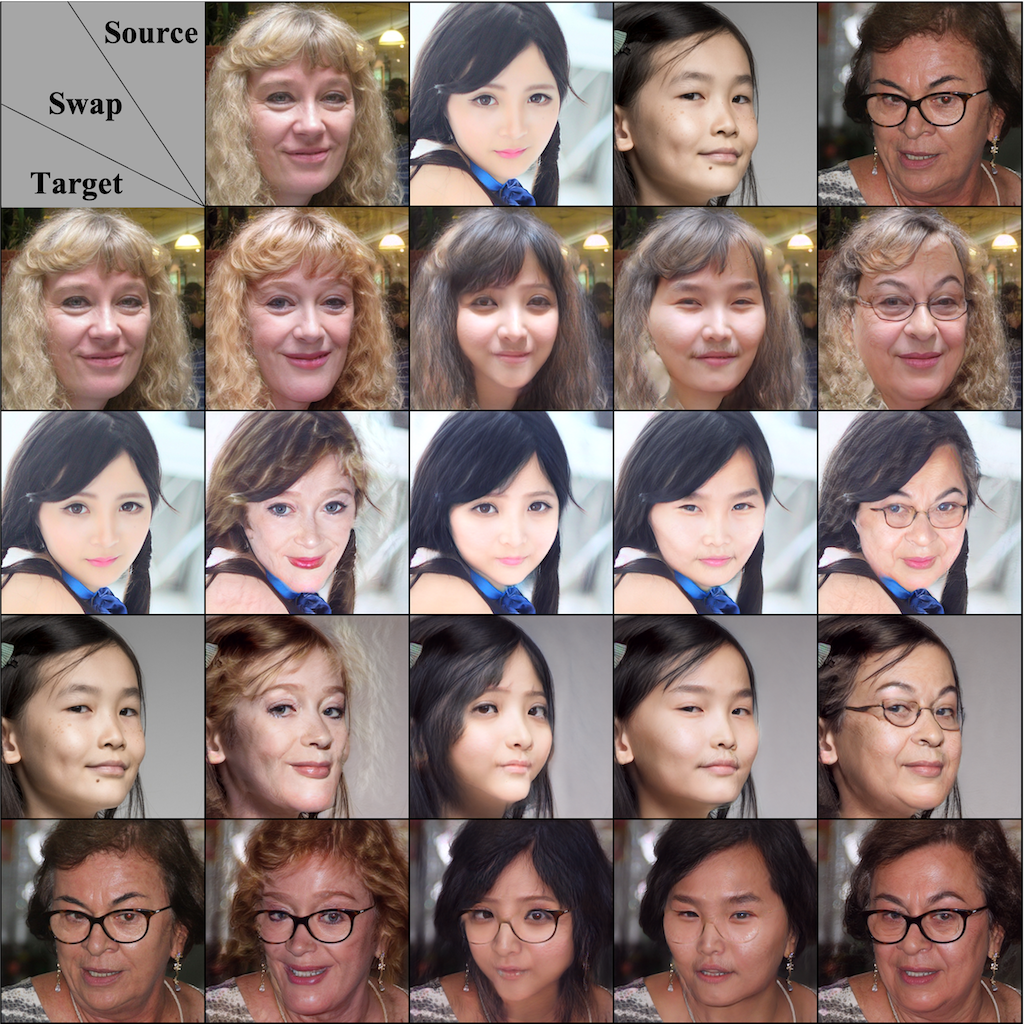}
    \caption{The results of Smooth-Swap on the FFHQ test split (uncurated). An active change of identity (e.g., row-1, column-2) is observed, but some artifacts can be also found when the source identity has a complicated hair pattern (column-1).}
    \label{fig:ffhq}
\end{figure}

We apply face-swapping on the FaceForensics++ dataset and compare the results with other models in Fig. \ref{fig:compare}. The figure shows that our Smooth-Swap model is more aggressive in changing identity, especially in face shape. For example, in the second and the fourth row, our swapped images show more round and grown chin shapes reflecting the characteristics of the source identity (more extreme cases can be found in Fig. \ref{fig:ffhq}); the images from the other models are mostly confined to textural change. Also, we can observe other identity-related attributes, such as skin tones or hair colors, are matched more to the source in our results, making the overall figure visually more close to the source. Fig.~\ref{fig:ffhq} and \ref{fig:wild} show the swapping results on FFHQ and face images in the wild (see appendix for more samples and discussion on the failure cases). 

The same trend can be seen from the quantitative results summarized in Table \ref{tab:synth}. In the table, Smooth-Swap shows good identity and shape scores (VGG, Arc, and Shp). While it is not as good in the other scores, it at least shows comparable numbers (not the worst at all times). Considering that a bypass model (not changing the identity at all) would achieve the best score in expression and pose scores, we emphasize that the overall competence is important here. Thus, we report an overall score in the last column---average of the metrics after standardizing each---where our model marks the best.

\begin{figure}
    \centering
    \includegraphics[width=0.90\columnwidth]{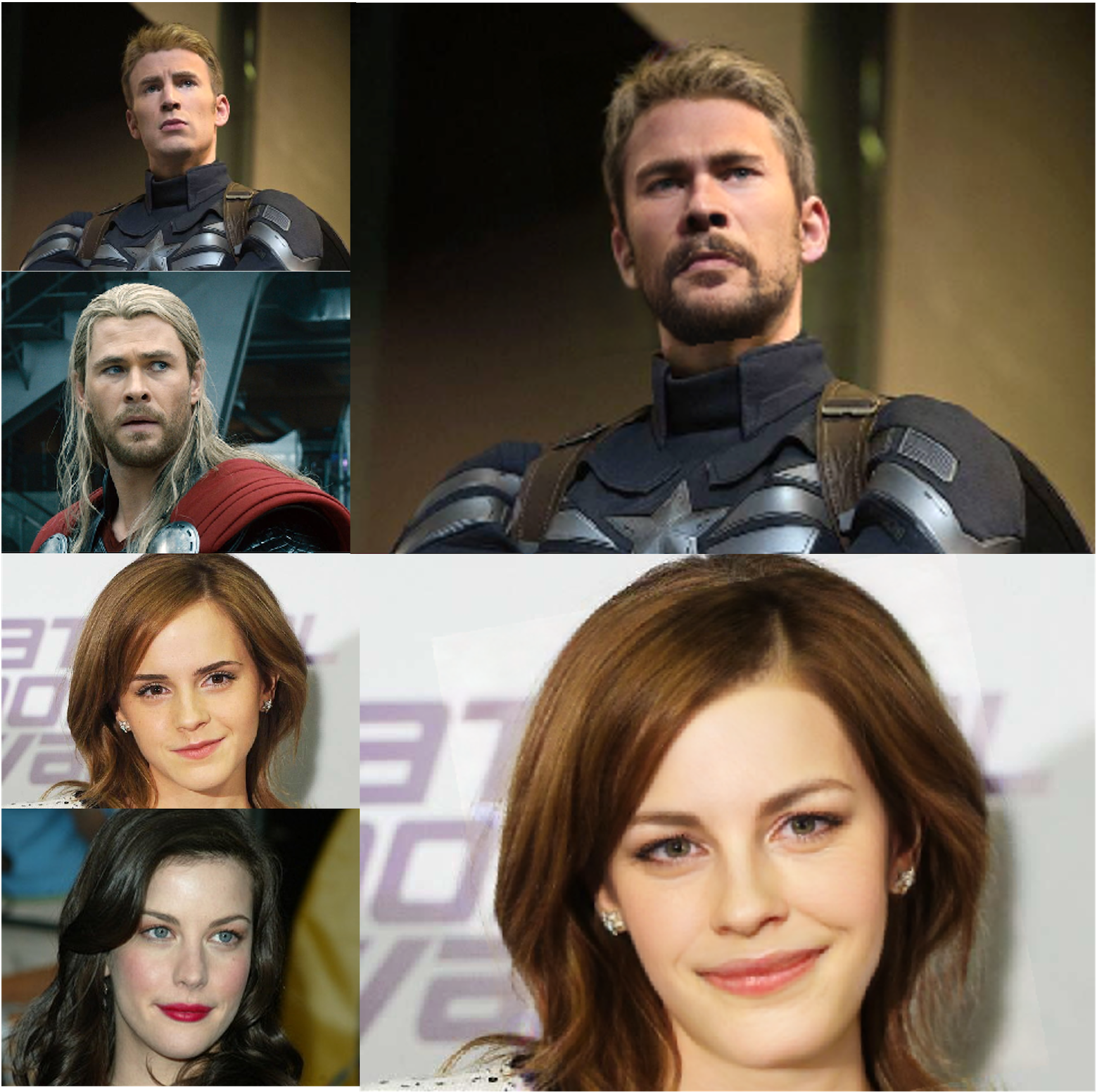}
    \caption{Face swapping results of Smooth-Swap on wild images. More samples are included in appendix.}
    \label{fig:wild}
\end{figure}

\subsection{Ablation Study on the Identity Embedder} \label{sec:ablation}

To see how our identity embedder makes a difference, we train our generator using ArcFace \cite{dengArcFaceAdditiveAngular2019} as well. As seen from the lower part of Table \ref{tab:synth}, the models using ArcFace perform worse in most of the metrics.

\begin{figure}
    \centering
    \includegraphics[width=\columnwidth]{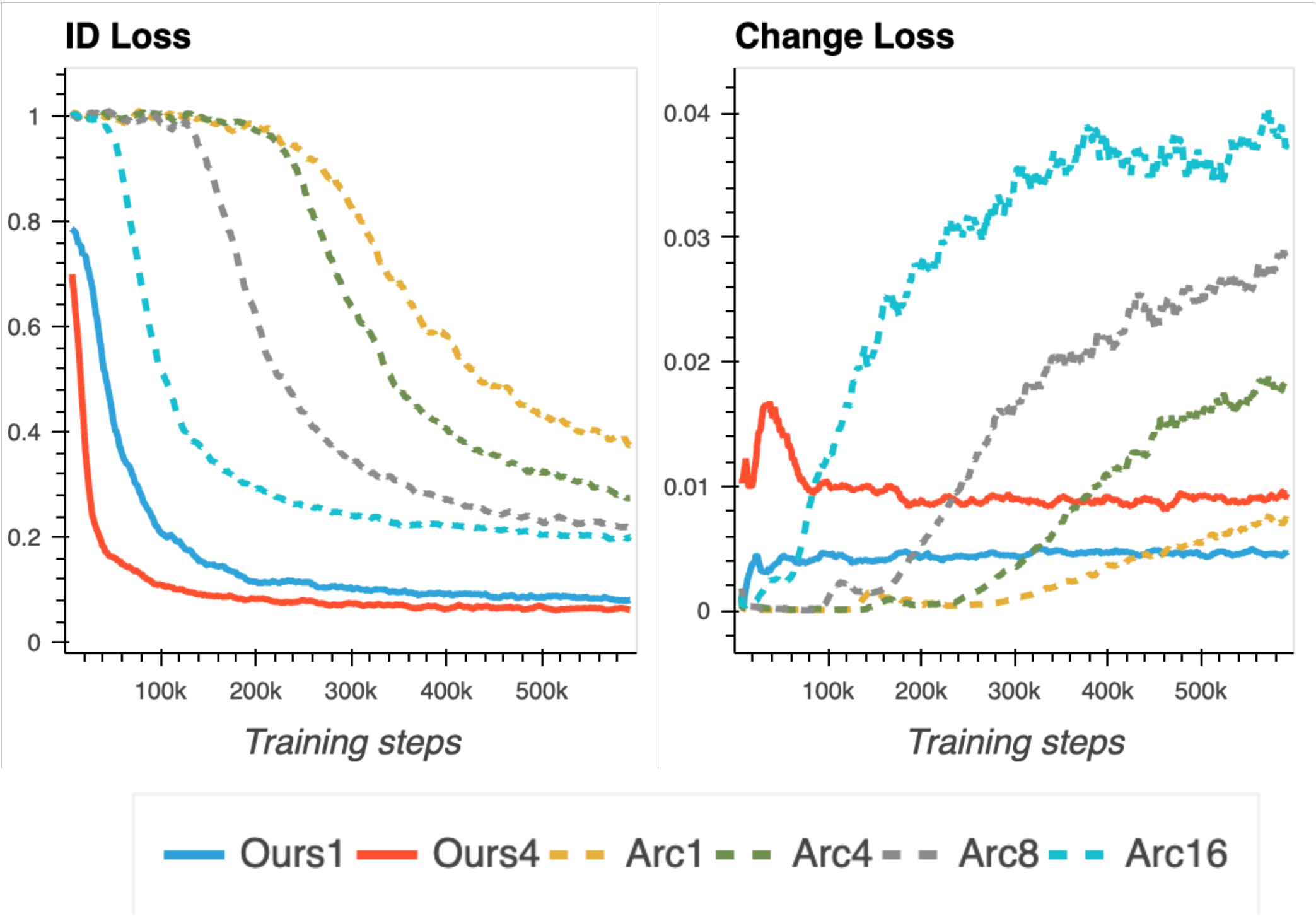}
    \caption{Ablation study of identity embedding model---Ours (solid) versus ArcFace (dashed) \cite{dengArcFaceAdditiveAngular2019}. The number next to the model name indicates the identity-loss weight, $\lambda_{id}$, used for training. It can be seen that the model learns to change identity much faster with our embedder while being stable in the change loss. See Sec. \ref{sec:ablation} for the discussion. 
    }
    \label{fig:emb_vs_loss}
\end{figure}

More importantly, we observe that our embedder enables faster and stable training. In Fig. \ref{fig:emb_vs_loss}, the left graph shows that the identity loss of our model converges faster compared with the one using ArcFace. Note this is not due to the scales or the choice of $\lambda_{id}$, since Arc16, which has a similar rate of identity-loss drop, shows a significantly worse curve for the change loss.

\begin{figure*}
    \centering
    \includegraphics[width=\textwidth]{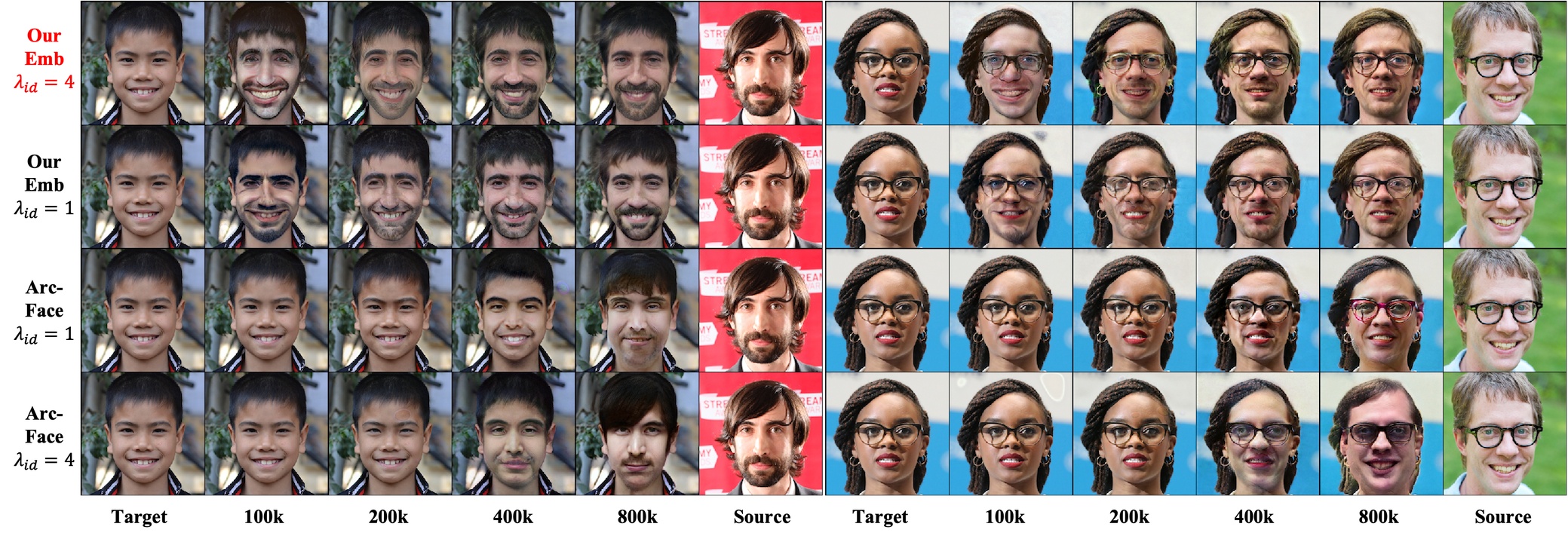}
    \caption{The progression of model training with different identity embedding models and loss weighting ($\lambda_{id}$); the generator architecture is fixed to ours. The models with ArcFace embedder \cite{dengArcFaceAdditiveAngular2019} shows slow training, making little identity change until being trained for 400k steps. On the other hand, the models with our embedder show identity change at as early as 100k steps. See Sec. \ref{sec:ablation}.}
    \label{fig:emb_progress}
\end{figure*}

The same trend can be seen in Fig. \ref{fig:emb_progress}. When paired with ArcFace embedder, the models show slow training, rarely changing identity until 400k training steps. In contrast, the models with our embedder begins to change the identity as early as 100k steps, and the overall score at 400k steps (50\% training) is already better than HifiFace (Table \ref{tab:synth}).

\begin{figure*}
    \centering
    \includegraphics[width=\textwidth]{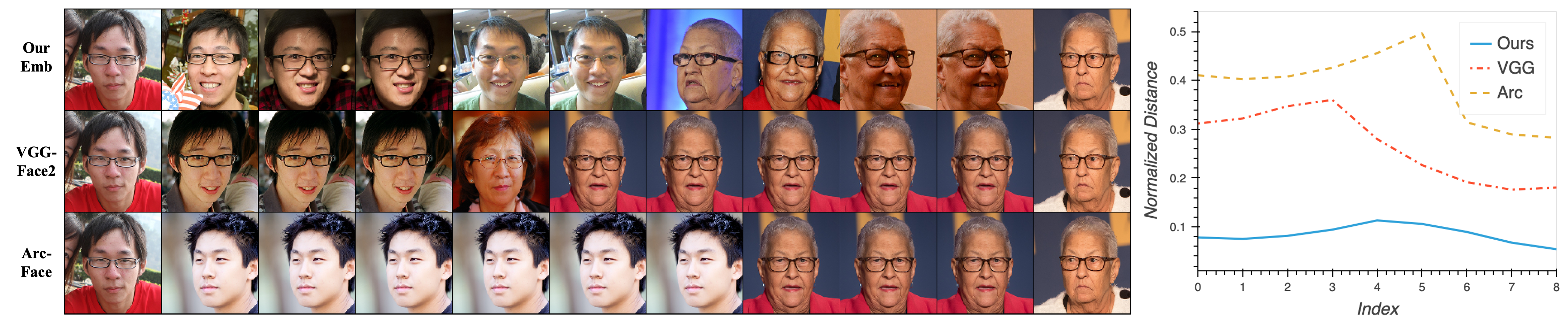}
    \caption{Inspection of the smoothness of embedders via interpolation. For two randomly picked images from the FFHQ test split (the leftmost and the rightmost), we compute the interpolations in the embedding space. For each of the nine interpolating points, we retrieve the closest images (compared in the embedding space) from the train split. Our embedder tends to show continuously changing identities, whereas others show repeating identities, implying non-smoothness of the space. The graph on the right shows our embedder distributes the identities more uniformly. The distances are normalized by the average of 4k random pairs for each embedder. See Sec. \ref{sec:id_emb_perf}.}
    \label{fig:emb_interp}
\end{figure*}

\subsection{Identity Embedding Performance} \label{sec:id_emb_perf}
The advantage expected from our identity embedder is the smoothness; in particular, smooth change of identities along the interpolation curve as shown in Fig. \ref{fig:why_smooth_emb}. To quantitatively evaluate this, we devised a smoothness score and compared with other baseline embedders. 
$$d_{smooth}:=E_{x_{A},x_{B}\sim p(x)}\left[\frac{\Vert Slerp(z_{A},z_{B};r)-z_{C}\Vert}{\Vert z_{A}-z_{B}\Vert}\right].$$
The score measures the (normalized) gap between the average point of the two identity embedding vectors, $Slerp(z_A, z_B; r)$, and the closest valid embedding to it, $z_C$ (here, $r$ is an averaging ratio). If the embedding space is smooth, this gap has to be small. 

The notion of valid embedding is subject to the settings. When measured using samples, $x_A$ and $x_B$ are samples from the FFHQ dataset $D_{test}$, and $z_C=f_{emb}(x_C)$ where $x_C=\arg\min_{x\in D_{train}} \Vert Slerp - f_{emb}(x) \Vert$. When measured using GAN, $x_A=g(y_A)$ and $x_B=g(y_B)$ are samples generated from a pretrained StyleGAN2 \cite{karrasAnalyzingImprovingImage2020}, where $z_C=f_{emb}(g(Lerp(y_A,y_B;r)))$ ($g$ is the generator, and $y$'s are the latent codes). 

As seen from Table \ref{tab:embedder}, our model shows substantially better smoothness while maintaining comparable verification performance with ArcFace and VGGFace2. Note LFW \cite{LFWTech} is one of the standard benchmark dataset for verification; VCHQ is a dataset we derived from VoxCeleb \cite{nagraniVoxcelebLargescaleSpeaker2019} (see appendix for the details). 

The same trend is also qualitatively confirmed in Fig. \ref{fig:emb_interp}. The figure shows the retrieved $x_C$ images for each of the interpolating points ($r\in[0.1, \cdots, 0.9]$). Our embedder tends to change smoothly while moving along the interpolation curve; others tend to stick with the same identities repeatedly. To quantify this, we compute the number of unique images for each interpolation (the lower, the more repetition, and the worse). Summarizing the results from 64 sample pairs, the numbers were 5.13$\pm$1.18 (Ours), 4.42$\pm$1.41 (VGGFace2), and 4.25$\pm$1.33 (ArcFace).

\begin{table}[]
    \centering
{\footnotesize
\begin{tabular}{cccc|ccc}
\toprule 
 & \multicolumn{2}{c}{\textbf{\footnotesize{}$d_{smooth}$ w/smp}} & \multicolumn{1}{c}{\textbf{\footnotesize{}w/GAN}} & \multicolumn{3}{c}{\textbf{\footnotesize{}Verification AUC}}\tabularnewline
\midrule 
 & \textbf{\footnotesize{}r=0.25} & \textbf{\footnotesize{}r=0.5} & \textbf{\footnotesize{}r=0.5} & \textbf{\footnotesize{}VCHQ} & \textbf{\footnotesize{}VGG2} & \textbf{\footnotesize{}LFW}\tabularnewline
\midrule
\midrule 
\textbf{\footnotesize{}CE-Lin} & {\footnotesize{}0.333} & {\footnotesize{}0.354} & {\footnotesize{}0.797} & {\footnotesize{}0.939} & {\footnotesize{}0.994} & {\footnotesize{}1.000}\tabularnewline
\midrule 
\textbf{\footnotesize{}CE-Arc} & {\footnotesize{}0.404} & {\footnotesize{}0.430} & {\footnotesize{}0.914} & {\footnotesize{}0.925} & {\footnotesize{}0.997} & {\footnotesize{}0.998}\tabularnewline
\midrule 
\textbf{\footnotesize{}ArcFace} & {\footnotesize{}0.360} & {\footnotesize{}0.380} & {\footnotesize{}0.802} & {\footnotesize{}-} & {\footnotesize{}-} & {\footnotesize{}0.995}\tabularnewline
\midrule 
\textbf{\footnotesize{}Ours} & \textbf{\footnotesize{}0.116} & \textbf{\footnotesize{}0.135} & \textbf{\footnotesize{}0.671} & {\footnotesize{}0.956} & {\footnotesize{}0.994} & {\footnotesize{}0.999}\tabularnewline
\bottomrule
\end{tabular}

}
    \caption{Scores of the embedder models. Our model shows far better smoothness scores, maintaining comparable verification scores. CE-Lin and CE-Arc are reproduced versions of VGGFace2 \cite{caoVGGFace2DatasetRecognising2018} and ArcFace \cite{dengArcFaceAdditiveAngular2019}, trained from FFHQ-aligned VGGFace2 dataset. ArcFace is the original model provided in \cite{dengArcFaceAdditiveAngular2019}, trained from a larger dataset with a different alignment.}
    \label{tab:embedder}
\end{table}

\section{Conclusion}
We introduced Smooth-Swap, a new face-swapping model generating high-quality swap images with active change of face shape. While existing models use handcrafted modules to tackle the difficulty, our model stays with the simplest architecture and considers smooth identity embedding instead. By taking this data-driven approach with minimal inductive bias, we observed that Smooth-Swap can achieve the best overall scores with fast convergence.

We believe this study can open up opportunity for tackling more challenging face-swapping problems by reducing the complexity considerably. With reduced effort for balancing the components and reduced memory usage, one could consider an expanded problem scope, such as modeling face-swapping on videos in an end-to-end manner. A downside of our current model in that regard is some performance drop in preserving the pose and expression. However, we suppose a simple fine-tuning or different hyperparameter choice would be sufficient to meet the goal.

\paragraph{Potential Negative Societal Impact} 
Face-swapping models, known as Deepfake to the public, have been maliciously used in making serious negative impacts (e.g., spread of fake news). Nonetheless, we believe studying on these models is important and necessary because deep understanding on them could set a good starting point for developing high-quality Deepfake detection algorithms \cite{rosslerFaceForensicsLearningDetect2019}. We remark that they also have positive applications, including anonymization for privacy protection and creating new characters without heavy CGI techniques. 

\paragraph{Acknowledgement} {\footnotesize This work was partly supported by IITP (2015-0-00310/20\%, 2018-0-00622/15\%, 2019-0-01371/20\%, 2021-0-02068/15\%, 2021-0-01343/15\%) grants, and CARAI (UD190031RD/15\%) grant by DAPA and ADD.}

{\small
\bibliographystyle{ieee_fullname}
\bibliography{face-swap}

\begin{thebibliography}{10}\itemsep=-1pt

\bibitem{DeepFakesHttpsGithub2021}
{{DeepFakes}} ({{https://github.com/deepfakes/faceswap)}}, Nov. 2021.

\bibitem{FaceSwapHttpsGithub2021}
{{FaceSwap}} ({{https://github.com/MarekKowalski/FaceSwap)}}, Nov. 2021.

\bibitem{baoOpenSetIdentityPreserving2018b}
Jianmin Bao, Dong Chen, Fang Wen, Houqiang Li, and Gang Hua.
\newblock Towards {{Open-Set Identity Preserving Face Synthesis}}.
\newblock In {\em 2018 {{IEEE}}/{{CVF Conference}} on {{Computer Vision}} and
  {{Pattern Recognition}}}, pages 6713--6722, {Salt Lake City, UT, USA}, June
  2018. {IEEE}.

\bibitem{blanzMorphableModelSynthesis1999}
Volker Blanz and Thomas Vetter.
\newblock A morphable model for the synthesis of {{3D}} faces.
\newblock In {\em Proceedings of the 26th Annual Conference on {{Computer}}
  Graphics and Interactive Techniques}, {{SIGGRAPH}} '99, pages 187--194,
  {USA}, July 1999. {ACM Press/Addison-Wesley Publishing Co.}

\bibitem{brockLargeScaleGAN2019}
Andrew Brock, Jeff Donahue, and Karen Simonyan.
\newblock Large {{Scale GAN Training}} for {{High Fidelity Natural Image
  Synthesis}}.
\newblock In {\em International Conference on Learning Representations}, 2019.

\bibitem{caoVGGFace2DatasetRecognising2018}
Qiong Cao, Li Shen, Weidi Xie, Omkar~M Parkhi, and Andrew Zisserman.
\newblock Vggface2: {{A}} dataset for recognising faces across pose and age.
\newblock In {\em 2018 13th {{IEEE}} International Conference on Automatic Face
  \& Gesture Recognition ({{FG}} 2018)}, pages 67--74. {IEEE}, 2018.

\bibitem{chenSimSwapEfficientFramework2020}
Renwang Chen, Xuanhong Chen, B. Ni, and Yanhao Ge.
\newblock {{SimSwap}}: {{An Efficient Framework For High Fidelity Face
  Swapping}}.
\newblock {\em ACM Multimedia}, 2020.

\bibitem{dengArcFaceAdditiveAngular2019}
Jiankang Deng, Jia Guo, Niannan Xue, and Stefanos Zafeiriou.
\newblock {{ArcFace}}: {{Additive Angular Margin Loss}} for {{Deep Face
  Recognition}}.
\newblock In {\em Proceedings of the {{IEEE}}/{{CVF}} Conference on Computer
  Vision and Pattern Recognition}, Feb. 2019.

\bibitem{gaoInformationBottleneckDisentanglement2021}
Gege Gao, Huaibo Huang, Chaoyou Fu, Zhaoyang Li, and Ran He.
\newblock Information {{Bottleneck Disentanglement}} for {{Identity Swapping}}.
\newblock In {\em 2021 {{IEEE}}/{{CVF Conference}} on {{Computer Vision}} and
  {{Pattern Recognition}} ({{CVPR}})}, pages 3403--3412, {Nashville, TN, USA},
  June 2021. {IEEE}.

\bibitem{goodfellowGenerativeAdversarialNetworks2014}
Ian~J. Goodfellow, Jean {Pouget-Abadie}, Mehdi Mirza, Bing Xu, David
  {Warde-Farley}, Sherjil Ozair, Aaron Courville, and Yoshua Bengio.
\newblock Generative {{Adversarial Networks}}.
\newblock {\em Advances in Neural Information Processing Systems}, 27, 2014.

\bibitem{heDeepResidualLearning2016}
Kaiming He, Xiangyu Zhang, Shaoqing Ren, and Jian Sun.
\newblock Deep {{Residual Learning}} for {{Image Recognition}}.
\newblock In {\em 2016 {{IEEE Conference}} on {{Computer Vision}} and {{Pattern
  Recognition}} ({{CVPR}})}, pages 770--778, {Las Vegas, NV, USA}, June 2016.
  {IEEE}.

\bibitem{LFWTech}
Gary~B. Huang, Manu Ramesh, Tamara Berg, and Erik {Learned-Miller}.
\newblock Labeled faces in the wild: {{A}} database for studying face
  recognition in unconstrained environments.
\newblock Technical Report 07-49, {University of Massachusetts, Amherst}, Oct.
  2007.

\bibitem{karrasTrainingGenerativeAdversarial2020}
Tero Karras, Miika Aittala, Janne Hellsten, Samuli Laine, Jaakko Lehtinen, and
  Timo Aila.
\newblock Training {{Generative Adversarial Networks}} with {{Limited Data}}.
\newblock {\em Advances in Neural Information Processing Systems},
  33:12104--12114, 2020.

\bibitem{karrasStyleBasedGeneratorArchitecture2019}
Tero Karras, Samuli Laine, and Timo Aila.
\newblock A {{Style-Based Generator Architecture}} for {{Generative Adversarial
  Networks}}.
\newblock In {\em Proceedings of the {{IEEE}}/{{CVF}} Conference on Computer
  Vision and Pattern Recognition ({{CVPR}})}, June 2019.

\bibitem{karrasAnalyzingImprovingImage2020}
Tero Karras, Samuli Laine, Miika Aittala, Janne Hellsten, Jaakko Lehtinen, and
  Timo Aila.
\newblock Analyzing and {{Improving}} the {{Image Quality}} of {{StyleGAN}}.
\newblock In {\em Proceedings of the {{IEEE}}/{{CVF}} Conference on Computer
  Vision and Pattern Recognition}, pages 8110--8119, 2020.
\newblock StyleGAN2.

\bibitem{khoslaSupervisedContrastiveLearning2020}
Prannay Khosla, Piotr Teterwak, Chen Wang, Aaron Sarna, Yonglong Tian, Phillip
  Isola, Aaron Maschinot, Ce Liu, and Dilip Krishnan.
\newblock Supervised {{Contrastive Learning}}.
\newblock {\em Advances in Neural Information Processing Systems},
  33:18661--18673, 2020.

\bibitem{kingmaAdamMethodStochastic2015}
Diederik~P. Kingma and Jimmy Ba.
\newblock Adam: {{A Method}} for {{Stochastic Optimization}}.
\newblock In {\em {{ICLR}} (Poster)}, 2015.

\bibitem{liFaceShifterHighFidelity2019}
Lingzhi Li, Jianmin Bao, Hao Yang, Dong Chen, and Fang Wen.
\newblock {{FaceShifter}}: {{Towards High Fidelity And Occlusion Aware Face
  Swapping}}.
\newblock {\em arXiv:1912.13457 [cs]}, Dec. 2019.

\bibitem{Li_2020_CVPR}
Lingzhi Li, Jianmin Bao, Hao Yang, Dong Chen, and Fang Wen.
\newblock Advancing high fidelity identity swapping for forgery detection.
\newblock In {\em Proceedings of the {{IEEE}}/{{CVF}} Conference on Computer
  Vision and Pattern Recognition ({{CVPR}})}, June 2020.

\bibitem{meschederWhichTrainingMethods2018a}
Lars Mescheder, Andreas Geiger, and Sebastian Nowozin.
\newblock Which {{Training Methods}} for {{GANs}} do actually {{Converge}}?
\newblock In {\em International Conference on Machine Learning}, pages
  3481--3490. {PMLR}, 2018.

\bibitem{nagraniVoxcelebLargescaleSpeaker2019}
Arsha Nagrani, Joon~Son Chung, Weidi Xie, and Andrew Zisserman.
\newblock Voxceleb: {{Large-scale}} speaker verification in the wild.
\newblock {\em Computer Science and Language}, 2019.

\bibitem{naruniecHighResolutionNeuralFace2020}
Jacek Naruniec, Leonhard Helminger, Christopher Schroers, and Romann~M Weber.
\newblock High-{{Resolution Neural Face Swapping}} for {{Visual Effects}}.
\newblock In {\em Computer Graphics Forum}, volume~39, pages 173--184. {Wiley
  Online Library}, 2020.

\bibitem{natsumeFSNetIdentityAwareGenerative2018}
Ryota Natsume, Tatsuya Yatagawa, and Shigeo Morishima.
\newblock {{FSNet}}: {{An Identity-Aware Generative Model}} for {{Image-based
  Face Swapping}}.
\newblock In {\em Asian Conference on Computer Vision}, pages 117--132.
  {Springer}, 2018.

\bibitem{natsumeRSGANFaceSwapping2018}
Ryota Natsume, Tatsuya Yatagawa, and Shigeo Morishima.
\newblock {{RSGAN}}: {{Face Swapping}} and {{Editing}} using {{Face}} and
  {{Hair Representation}} in {{Latent Spaces}}.
\newblock In {\em {{ACM SIGGRAPH}} 2018 Posters}, {{SIGGRAPH}} '18, {New York,
  NY, USA}, 2018. {Association for Computing Machinery}.

\bibitem{nirkinFSGANSubjectAgnostic2019}
Yuval Nirkin, Yosi Keller, and Tal Hassner.
\newblock {{FSGAN}}: {{Subject Agnostic Face Swapping}} and {{Reenactment}}.
\newblock In {\em Proceedings of the {{IEEE}}/{{CVF}} International Conference
  on Computer Vision}, pages 7184--7193, 2019.

\bibitem{nirkinFaceSegmentationFace2018}
Yuval Nirkin, Iacopo Masi, Anh Tran~Tuan, Tal Hassner, and Gerard Medioni.
\newblock On {{Face Segmentation}}, {{Face Swapping}}, and {{Face Perception}}.
\newblock In {\em 2018 13th {{IEEE International Conference}} on {{Automatic
  Face}} \& {{Gesture Recognition}} ({{FG}} 2018)}, pages 98--105, {Xi'an}, May
  2018. {IEEE}.

\bibitem{ronnebergerUNetConvolutionalNetworks2015}
Olaf Ronneberger, Philipp Fischer, and Thomas Brox.
\newblock U-{{Net}}: {{Convolutional Networks}} for {{Biomedical Image
  Segmentation}}.
\newblock In {\em International {{Conference}} on {{Medical}} Image Computing
  and Computer-Assisted Intervention}, pages 234--241. {Springer}, 2015.

\bibitem{rosslerFaceForensicsLearningDetect2019}
Andreas Rossler, Davide Cozzolino, Luisa Verdoliva, Christian Riess, Justus
  Thies, and Matthias Niessner.
\newblock {{FaceForensics}}++: {{Learning}} to {{Detect Manipulated Facial
  Images}}.
\newblock In {\em 2019 {{IEEE}}/{{CVF International Conference}} on {{Computer
  Vision}} ({{ICCV}})}, pages 1--11, {Seoul, Korea (South)}, Oct. 2019. {IEEE}.

\bibitem{sanyalLearningRegress3D2019}
Soubhik Sanyal, Timo Bolkart, Haiwen Feng, and Michael~J Black.
\newblock Learning to {{Regress 3D Face Shape}} and {{Expression}} from an
  {{Image}} without {{3D Supervision}}.
\newblock In {\em Proceedings of the {{IEEE}}/{{CVF}} Conference on Computer
  Vision and Pattern Recognition}, pages 7763--7772, 2019.

\bibitem{songScoreBasedGenerativeModeling2021}
Yang Song, Jascha {Sohl-Dickstein}, Diederik~P Kingma, Abhishek Kumar, Stefano
  Ermon, and Ben Poole.
\newblock Score-{{Based Generative Modeling}} through {{Stochastic Differential
  Equations}}.
\newblock In {\em International Conference on Learning Representations}, 2021.

\bibitem{thiesDeferredNeuralRendering2019}
Justus Thies, Michael Zollh{\"o}fer, and Matthias Nie{\ss}ner.
\newblock Deferred {{Neural Rendering}}: {{Image Synthesis}} using {{Neural
  Textures}}.
\newblock {\em ACM Transactions on Graphics (TOG)}, 38(4):1--12, 2019.
\newblock Comment: Video: https://youtu.be/z-pVip6WeyY SIGGRAPH 2019.

\bibitem{thiesFace2FaceRealtimeFace2016}
Justus Thies, Michael Zollhofer, Marc Stamminger, Christian Theobalt, and
  Matthias Nie{\ss}ner.
\newblock {{Face2Face}}: {{Real-time Face Capture}} and {{Reenactment}} of
  {{RGB Videos}}.
\newblock In {\em Proceedings of the {{IEEE}} Conference on Computer Vision and
  Pattern Recognition}, pages 2387--2395, 2016.

\bibitem{wangUnderstandingContrastiveRepresentation2020}
Tongzhou Wang and Phillip Isola.
\newblock Understanding {{Contrastive Representation Learning}} through
  {{Alignment}} and {{Uniformity}} on the {{Hypersphere}}.
\newblock In {\em International Conference on Machine Learning}, pages
  9929--9939. {PMLR}, 2020.

\bibitem{wangHifiFace3DShape2021}
Yuhan Wang, Xu Chen, Junwei Zhu, Wenqing Chu, Ying Tai, Chengjie Wang, Jilin
  Li, Yongjian Wu, Feiyue Huang, and Rongrong Ji.
\newblock {{HifiFace}}: {{3D Shape}} and {{Semantic Prior Guided High Fidelity
  Face Swapping}}.
\newblock In Zhi-Hua Zhou, editor, {\em Proceedings of the {{Thirtieth
  International Joint Conference}} on {{Artificial Intelligence}},
  {{IJCAI-21}}}, pages 1136--1142. {International Joint Conferences on
  Artificial Intelligence Organization}, Aug. 2021.

\bibitem{zhangUnreasonableEffectivenessDeep2018}
Richard Zhang, Phillip Isola, Alexei~A Efros, Eli Shechtman, and Oliver Wang.
\newblock The {{Unreasonable Effectiveness}} of {{Deep Features}} as a
  {{Perceptual Metric}}.
\newblock In {\em Proceedings of the {{IEEE}} Conference on Computer Vision and
  Pattern Recognition}, pages 586--595, 2018.

\bibitem{zhuOneShotFace2021}
Yuhao Zhu, Qi Li, Jian Wang, Cheng-Zhong Xu, and Zhenan Sun.
\newblock One {{Shot Face Swapping}} on {{Megapixels}}.
\newblock In {\em Proceedings of the {{IEEE}}/{{CVF}} Conference on Computer
  Vision and Pattern Recognition}, pages 4834--4844, 2021.

\end{thebibliography}
}

\clearpage
\appendix


\setcounter{figure}{0}    

\renewcommand{\thefigure}{A\arabic{figure}}
\renewcommand\thesection{\Alph{section}}

{\LARGE \textbf{Appendix}}

\section{Architecture Details}
\subsection{Identity Embedding Model: $f_{emb}^*$}
Our identity embedding model is based on ResNet-50 \cite{heDeepResidualLearning2016}, where we use a different head for contrastive learning as seen in Fig. \ref{appx-fig:arch}. Note the \textit{UnitNorm} in the final layer makes $z_{src}$ to be unit-length ($\Vert z_{src} \Vert=1$). The network invovles total 32.3M parameters.

\subsection{Swap-Image Generator: $f_{gen}$} \label{appx-sec:gen}
\begin{figure*}
    \centering
    \includegraphics[width=\textwidth]{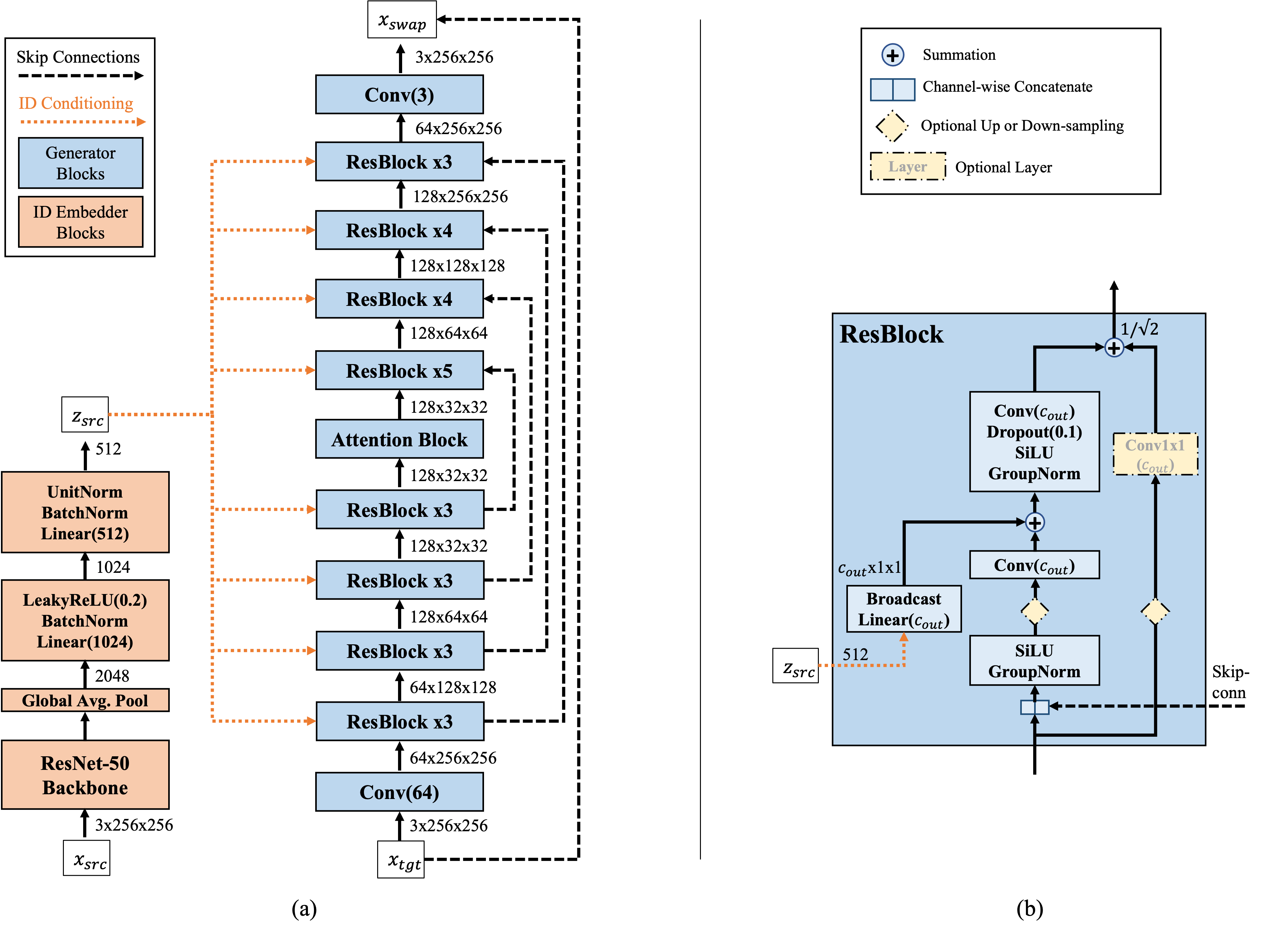}
    \caption[Detailed Architecture]{Detailed architecture of our Smooth-Swap model; both the identity embedder and the generator are shown. The intermediate feature-map dimensions are written in the order of (channels $\times$ height $\times$ width). `ResBlock x4' in (a) denotes that there are four residual sub-blocks connected sequentially; the structure of the sub-block (i.e., ResBlock) is detailed in (b). Note the multi-line text inside blocks should be read from the bottom (e.g., Linear(512) to BatchNorm to UnitNorm for the last embedder block). See Sec. \ref{appx-sec:gen}.}
    \label{appx-fig:arch}
\end{figure*}

Our generator architecture is mostly the same as NCSN++ \cite{songScoreBasedGenerativeModeling2021} except for the following three differences (as described in the main manuscript, Sec. 4.2): 1) we use half as many channels, 2) we use the identity embedding instead of the time embedding, and 3) we add an input-to-output skip connection. Fig. \ref{appx-fig:arch} (a) shows the detailed structure with dimensional information. The network involves total 9.8M parameters.

\paragraph{Up/Down Sampling \& Skip-Connections}
Note in each of the outer block containing multiple ResBlocks, the first ResBlock handles upsampling or downsampling (except for the ResBlock x5, where the second ResBlock handles upsampling). There are 13 skip connections in total ($13=3 \times 4 + 1$; $+1$ is the input-to-output skip), where the input to each of the ResBlock in the encoder part (before the Attention Block) is handed over to the decoder part (after the Attention Block). On the decoder side, the first three ResBlocks of each outer block get the skip-connections (except for the ResBlock x5, where the second through the fourth get the skip-connections).

\paragraph{Details on the ResBlocks of the Generator}
We describe some essential details of the \textit{ResBlocks} of the generator here. The complete information can be found in \cite{songScoreBasedGenerativeModeling2021}. 

The overall structure of ResBlock is not much different from the conventional design \cite{heDeepResidualLearning2016}. However, as shown in Fig. \ref{appx-fig:arch} (b), a structure for conditioning on the identity embedding vector $z_{src}$ is added (similar to \cite{brockLargeScaleGAN2019}). The conditioning is done by 1) projecting $z_{src}$ onto a $c_{out}$-dimensional vector, 2) spatially broadcasting the result, and 3) adding it to the intermediate output of the original path ($c_{out}$ is the number of output channels of the current block). When upsampling or downsampling is used, the optional components (denoted by yellow and dash-dotted outline) are also computed.

\paragraph{Throughput and FLOPs at Inference Time}
Smooth-Swap generator has much higher FLOPs (in MACs) than HifiFace \cite{wangHifiFace3DShape2021} (214.47G to 102.39G). However, it shows far better throughput (42.96 fps) than others (SimSwap \cite{chenSimSwapEfficientFramework2020}: 31.17, HifiFace \cite{wangHifiFace3DShape2021}: 25.29; FaceShifter \cite{liFaceShifterHighFidelity2019}: 22.34)\footnote{Test settings and values of other models are adopted from \cite{wangHifiFace3DShape2021}}. We speculate this is due to the simple and homogeneous architecture, which is advantageous for speed-up with GPU computing.

\subsection{Discrminiator: $f_{dis}$}
We use the same discriminator as the one used in StyleGAN2 \cite{karrasAnalyzingImprovingImage2020}. The network involves total 28.9M parameters.

\section{More Image Samples from Smooth-Swap}
We show extended sets of swapped-image samples from our Smooth-Swap model. The following three figures, Fig. \ref{appx-fig:ffpp}, \ref{appx-fig:ffhq}, and \ref{appx-fig:wild} present the results of the same experiments as Fig. 4, 5, and 6 in the main manuscript, but with different source and target pairs. Fig. \ref{appx-fig:met} shows the results for out-of-distribution cases, where oil paintings (Metfaces dataset \cite{karrasTrainingGenerativeAdversarial2020}) are used for swapping. Although the model is never trained on such images, the results are of decent quality, reflecting the characteristics of the source and the target with shape change.

\begin{figure}[!ht]
    \centering
    \includegraphics[width=\columnwidth]{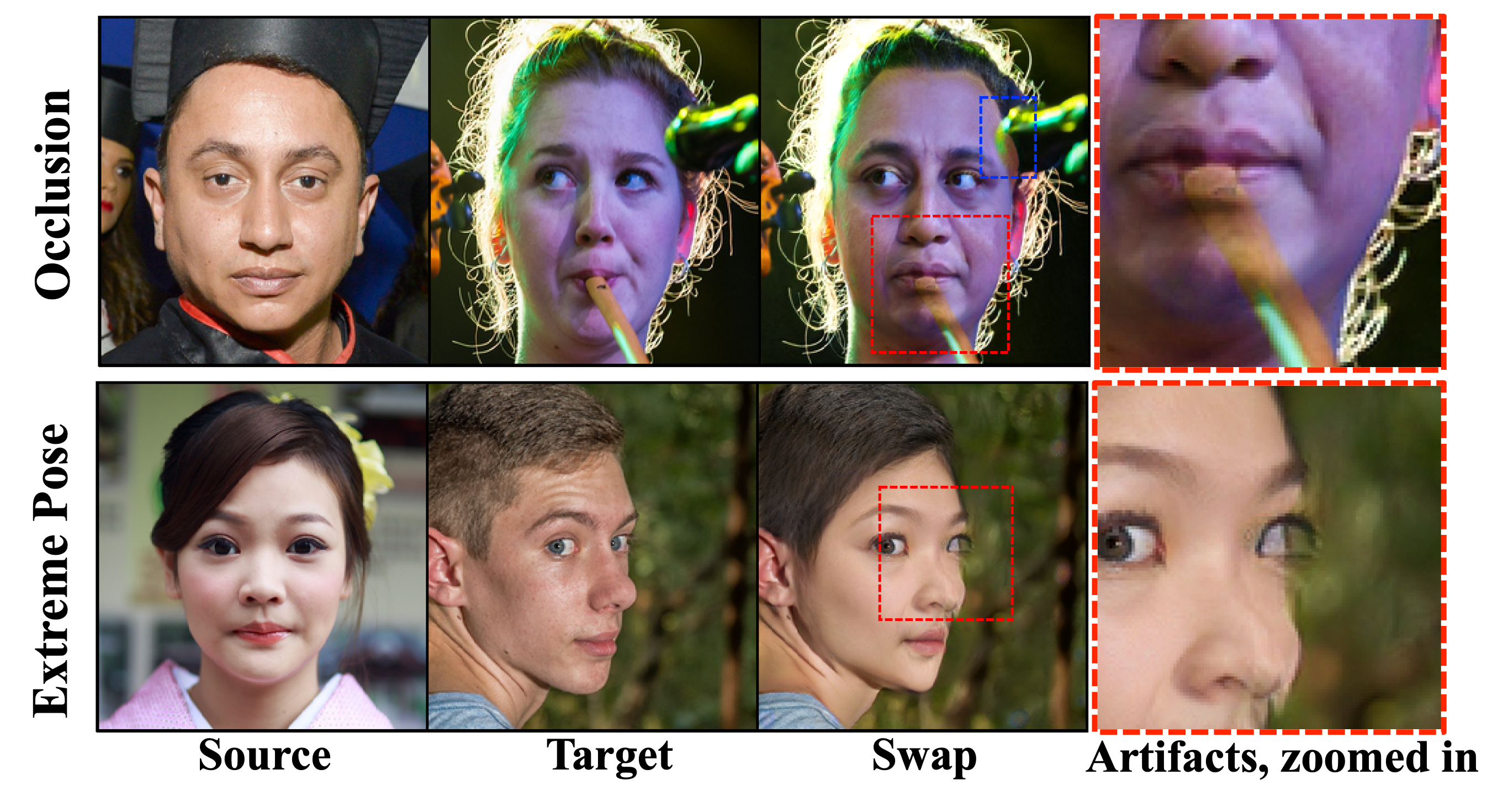}
    \caption[Failure Cases]{Some failure cases of Smooth-Swap. See Sec. \ref{appx-sec:limit}}
    \label{appx-fig:limit}
\end{figure}

\section{Extreme Cases and Limitations}
\label{appx-sec:limit}
We note that Smooth-Swap can fail when a target image involves occlusion or an extreme pose as shown in Fig. \ref{appx-fig:limit}. However, we believe each case can be handled by post-processing (e.g., HEAR-Net of \cite{liFaceShifterHighFidelity2019}) and supplying more extreme-pose examples for training.

\begin{figure*}
    \centering
    \includegraphics[width=\textwidth]{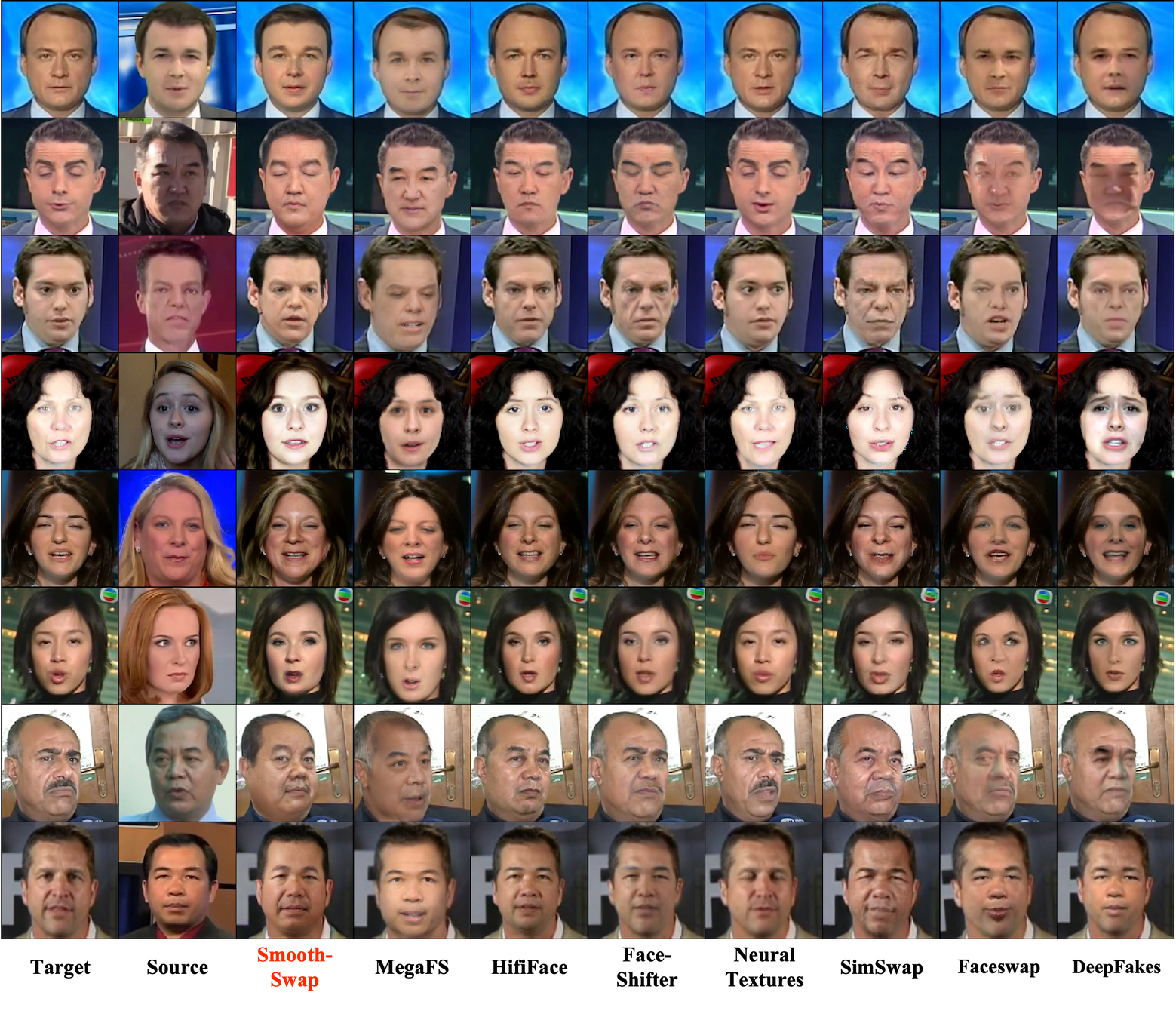}
    \caption[More FaceForensics++ Results]{Comparison of the face-swapping results of various models on the FaceForensics++ dataset \cite{rosslerFaceForensicsLearningDetect2019} (extension of Fig. 4 in the main manuscript)}
    \label{appx-fig:ffpp}
\end{figure*}

\begin{figure*}
    \centering
    \includegraphics[width=\textwidth]{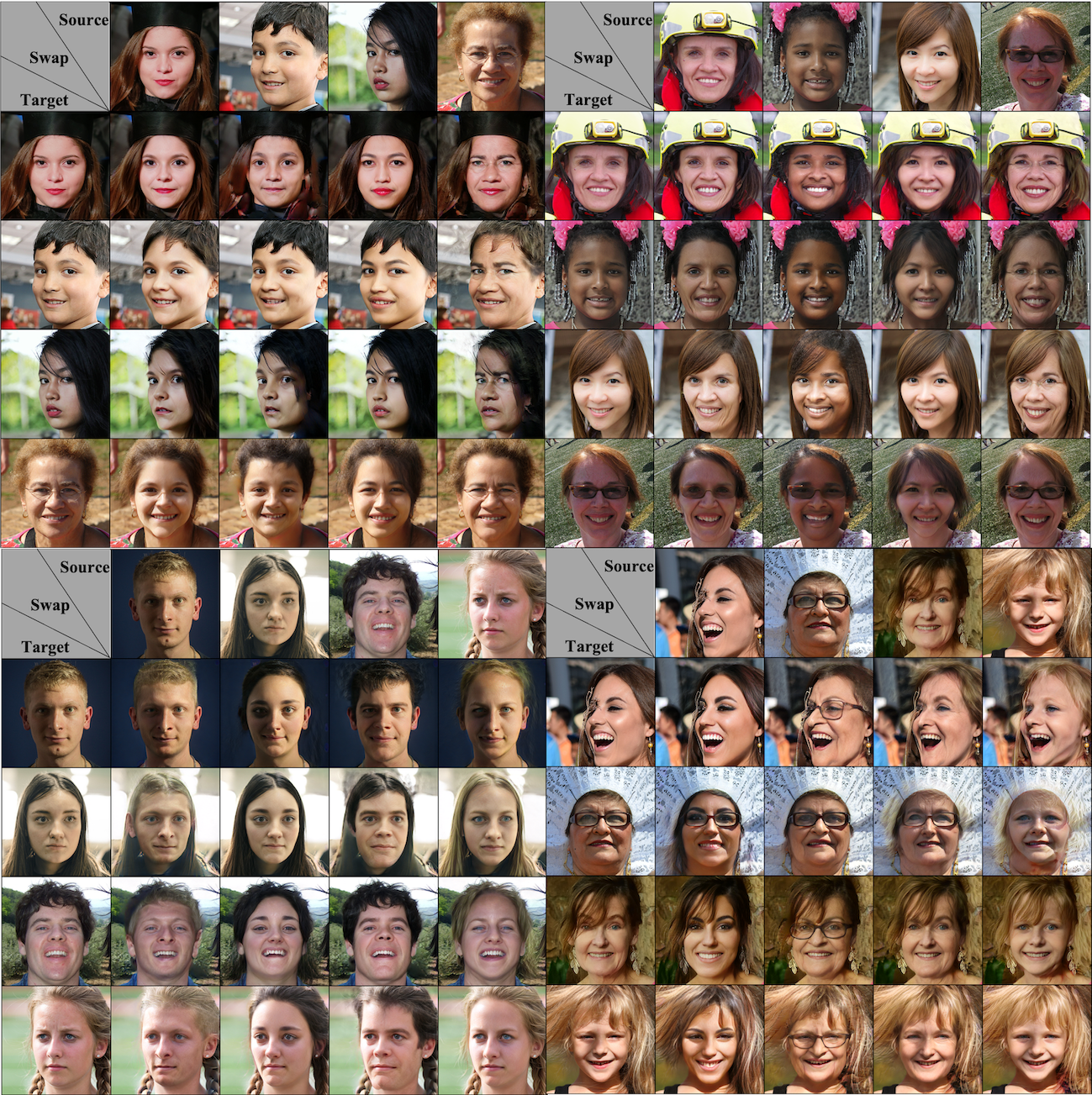}
    \caption[More FFHQ Results]{More results of Smooth-Swap on the FFHQ test split (extension of Fig. 5 in the main manuscript). Active change of identity is observed. However, in some cases where the source and the target have largely different face shapes (e.g., a child in the rightmost column in the lower-right block), artifacts are noticed. In real-world applications, such cases can be avoided by choosing the swapping pairs from a similar age range.}
    \label{appx-fig:ffhq}
\end{figure*}

\begin{figure*}
    \centering
    \includegraphics[width=\textwidth]{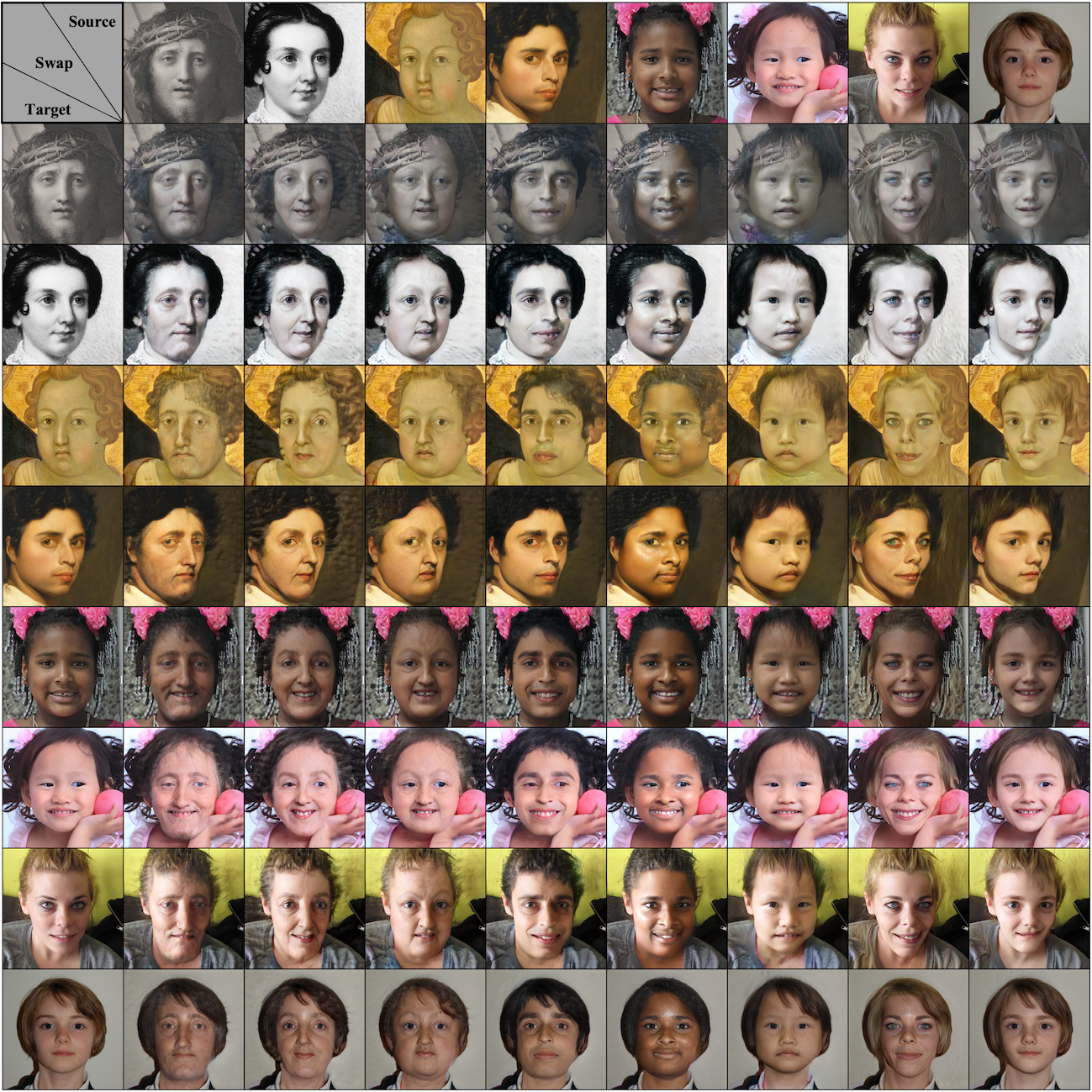}
    \caption[Face-Swapping on Metfaces]{Results of Smooth-Swap across the FFHQ test split and Metfaces \cite{karrasTrainingGenerativeAdversarial2020}. Even though the model is not trained on the oil paintings of Metfaces, it can still produce swapped images with a decent quality}
    \label{appx-fig:met}
\end{figure*}

\begin{figure*}
    \centering
    \includegraphics[width=\textwidth]{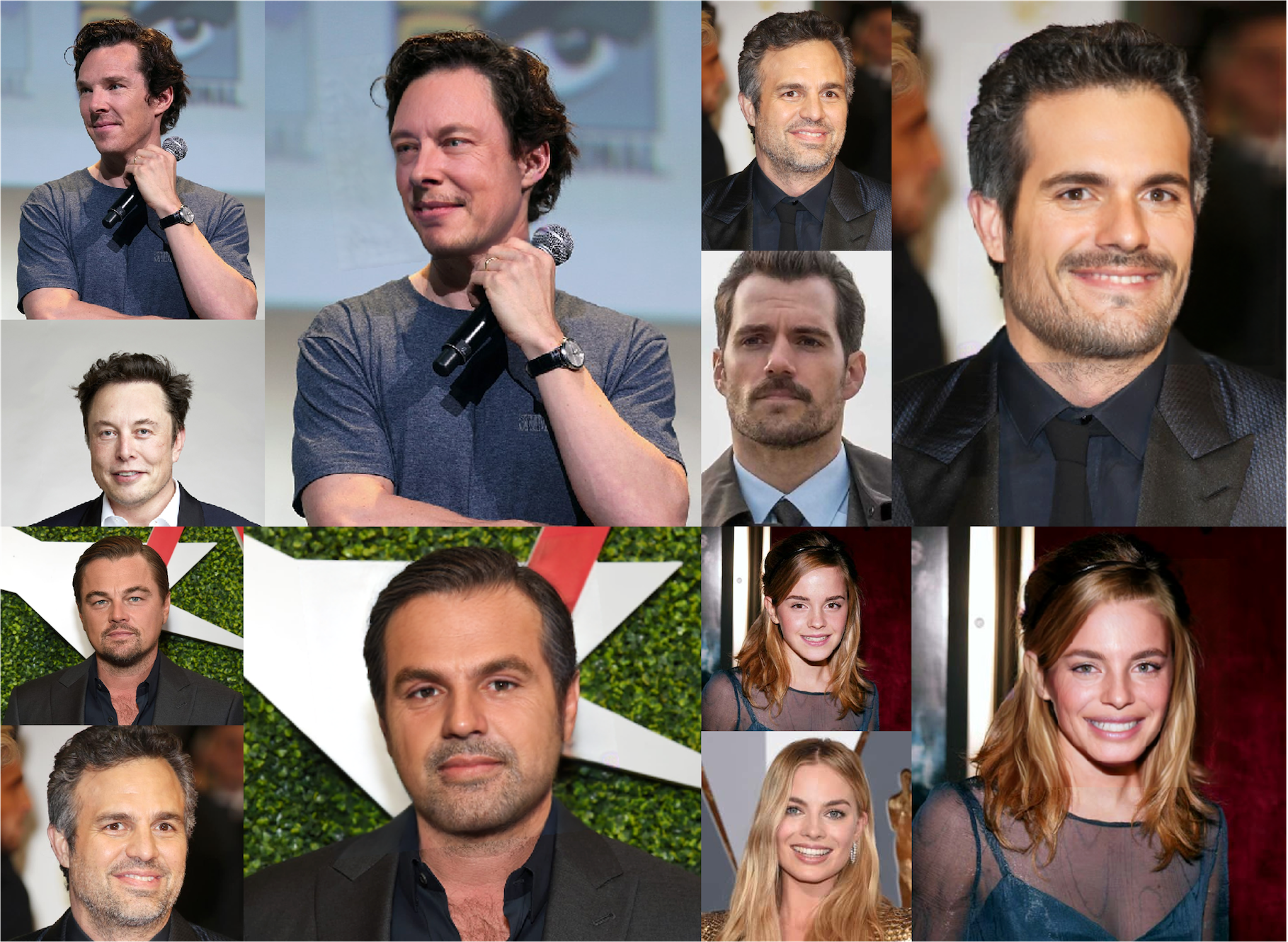}
    \caption[More Wild-image Results]{More face swapping results of Smooth-Swap on wild images (extension of Fig. 6 in the main manuscript).}
    \label{appx-fig:wild}
\end{figure*}

\end{document}